\documentclass[10pt,twocolumn,letterpaper]{article}

\usepackage{cvpr}              
\usepackage{float}
\usepackage{bm}
\usepackage{amsmath}
\definecolor{cvprblue}{rgb}{0.21,0.49,0.74}
\usepackage[pagebackref,breaklinks,colorlinks,allcolors=cvprblue]{hyperref}
\usepackage{enumerate}
\usepackage{multirow}
\usepackage{colortbl}
\usepackage{utfsym}
\usepackage{amssymb}
\usepackage{bbding}
\usepackage{xcolor} 
\usepackage{tabularx}
\usepackage{sidecap}
\usepackage{nicematrix}
\usepackage{algorithm}
\usepackage{algpseudocode}
\usepackage{amsmath}
\usepackage[misc]{ifsym}

\def\paperID{6909} 
\def\confName{CVPR}
\def\confYear{2025}

\title{Adv-CPG: A Customized Portrait Generation Framework with Facial Adversarial Attacks}

\author{
Junying Wang\textsuperscript{\rm 1} \ \ \ Hongyuan Zhang\textsuperscript{\rm 2,}\textsuperscript{\rm 3} \ \ \ Yuan Yuan\textsuperscript{\rm 1}\footnotemark[1]\\
\\
{\small \textsuperscript{\rm 1}School of Artificial Intelligence, OPtics and ElectroNics (iOPEN), Northwestern Polytechnical University, Xi'an, China} \\
{\small \textsuperscript{\rm 2}The University of Hong Kong, Hong Kong SAR, China}
{\small \textsuperscript{\rm 3}Institute of Artificial Intelligence (TeleAI), China Telecom, Shanghai, China}\\
{\tt\small  junyingwang959@163.com, hyzhang98@gmail.com, y.yuan.ieee@gmail.com}
}

\begin{document}
\twocolumn[{%
\renewcommand\twocolumn[1][]{#1}%
\maketitle
\begin{center}
    \vspace{-0.3cm}
    \centering
    \includegraphics[width=0.91\linewidth]{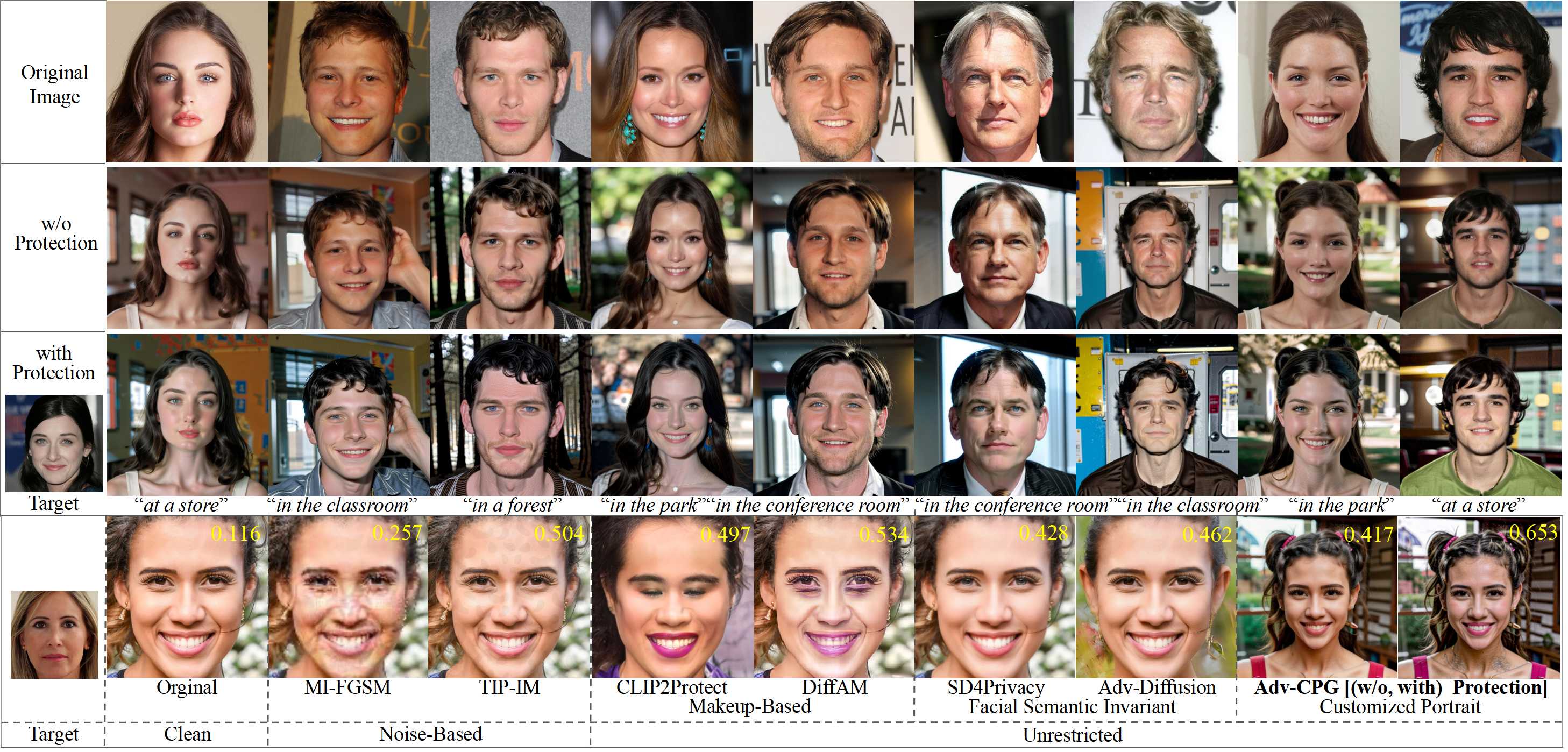}
    \vspace{-0.15cm}
    \captionof{figure}{The proposed Adv-CPG generates safe portraits that can deceive malicious face recognition systems. First row: original image. Second row: customized portrait based on the text prompt, without protection via Adv-CPG. Third row: customized portrait based on the scene text prompt, with protection via Adv-CPG. Fourth row: comparison with existing methods. The yellow number over each image: confidence score returned by Face++ when using the adversarial example for identity matching with the target (higher is better).}
    \label{fig:head}
\end{center}
}]

\maketitle

\renewcommand{\thefootnote}{\fnsymbol{footnote}} 
\footnotetext[1]{Corresponding author.} 

\begin{abstract}
Recent Customized Portrait Generation (CPG) methods, taking a facial image and a textual prompt as inputs, have attracted substantial attention. Although these methods generate high-fidelity portraits, they fail to prevent the generated portraits from being tracked and misused by malicious face recognition systems. To address this, this paper proposes a \textbf{C}ustomized \textbf{P}ortrait \textbf{G}eneration framework with facial \textbf{Adv}ersarial attacks (Adv-CPG). Specifically, to achieve facial privacy protection, we devise a lightweight local ID encryptor and an encryption enhancer. They implement progressive double-layer encryption protection by directly injecting the target identity and adding additional identity guidance, respectively. Furthermore, to accomplish fine-grained and personalized portrait generation, we develop a multi-modal image customizer capable of generating controlled fine-grained facial features. To the best of our knowledge, Adv-CPG is the first study that introduces facial adversarial attacks into CPG. Extensive experiments demonstrate the superiority of Adv-CPG, e.g.,
the average attack success rate of the proposed Adv-CPG is 28.1\% and 2.86\% higher compared to the SOTA noise-based attack methods and unconstrained attack methods, respectively.
\end{abstract}
\begin{table*}[!t]
    \centering
    \caption{Comparison of different facial privacy protection methods (noised-based vs. unrestricted) between unrestricted (stealthy), natural outputs, flexible text-guided adversaries, customized portraits, and fine-grained faces.}
    \label{tab:1}       
     \vspace{-0.2cm}
    \scalebox{0.65}{\begin{tabular}{c|cc|cccc|cc|c}
        \hline\noalign{\smallskip}
        \textbf{Method}  & \multicolumn{2}{c|}{\textbf{Noise-Based}} & \multicolumn{7}{c}{\textbf{Unrestricted}}\\
         & \multicolumn{2}{c|}{\textbf{}} & \multicolumn{4}{c}{\textbf{Makeup-Based}} &\multicolumn{2}{c}{\textbf{Facial Semantic Invariant}} & \textbf{Customized Portrait}\\
         &PGD~\cite{madry2017towards}   & TIP-IM~\cite{yang2021towards} & Aav-Makeup~\cite{yin2021adv}   & Clip2Protect~\cite{shamshad2023clip2protect}  &GIFT~\cite{li2024transferable} &DiffAM~\cite{sun2024diffam} &SD4Privacy~\cite{an2024sd4privacy} &Adv-Diffusion~\cite{liu2024adv} & Adv-CPG (Ours)\\
        \noalign{\smallskip}\hline\noalign{\smallskip}
        Stealthiness & &  &\usym{1F5F8} &\usym{1F5F8} &\usym{1F5F8} &\usym{1F5F8}& \usym{1F5F8}  &\usym{1F5F8} & \usym{1F5F8}\\
        Natural Outputs & & \Checkmark\kern-1.2ex\raisebox{1ex}{\rotatebox[origin=c]{125}{\textbf{--}}} &\usym{1F5F8} &\usym{1F5F8} &\usym{1F5F8}& \usym{1F5F8}& \usym{1F5F8}  & \usym{1F5F8} & \usym{1F5F8}\\
        Text Guided & & & &\usym{1F5F8} &  &\usym{1F5F8} &\usym{1F5F8} & & \cellcolor[RGB]{231,239,250}\usym{1F5F8}\\
        Customized Portrait &  &   & &  &   &  & & & \cellcolor[RGB]{231,239,250}\usym{1F5F8} \\
        Fine-Grained Face &  &  &  &   &  & &  & & \cellcolor[RGB]{231,239,250}\usym{1F5F8}\\
        \noalign{\smallskip}\hline
    \end{tabular}}
    \vspace{-0.4cm}
\end{table*}

\vspace{-0.7cm}
\section{Introduction}
\label{sec:intro}

Diffusion models \cite{rombach2022high, podell2023sdxl} have exhibited impressive superiority in the field of text-to-image (T2I) synthesis, especially for image customization tasks \cite{ guo2024pulid,kong2024anymaker}. Among these, Customized Portrait Generation (CPG) creates a portrait image that is visually consistent with the reference facial image, while simultaneously adhering to additional prompts. Early CPG approaches \cite{gal2022image, hulora,ruiz2024hyperdreambooth} achieve admirable high-fidelity results but require fine-tuning at test time, which limits their practicability. Recent test-time tuning-free methods \cite{li2024photomaker,wang2024instantid,xiao2024fastcomposer,ye2023ip} typically inject visual features of the reference face and textual features from language prompts, leading to consistent and high-fidelity portrait generation with the promise of large-scale deployment. However, to the best of our knowledge, \textit{current CPG methods lack facial security mechanisms to protect the generated high-fidelity portraits from being tracked and abused by malicious face recognition (FR) systems}.

Meanwhile, although existing noise-based \cite{,dong2018boosting, yang2021towards} and unrestricted \cite{sharif2016accessorize, jia2022adv} adversarial attack methods have achieved satisfactory facial privacy protection, they merely operate on the original face, \textit{which meets the protection needs of users but falls short of their portrait customization requirements}. In the current social media-enriched era, a safe and attractive personalized portrait is highly coveted by many people.



\begin{figure*}[t]
  \centering
   \includegraphics[width=1.0\linewidth]{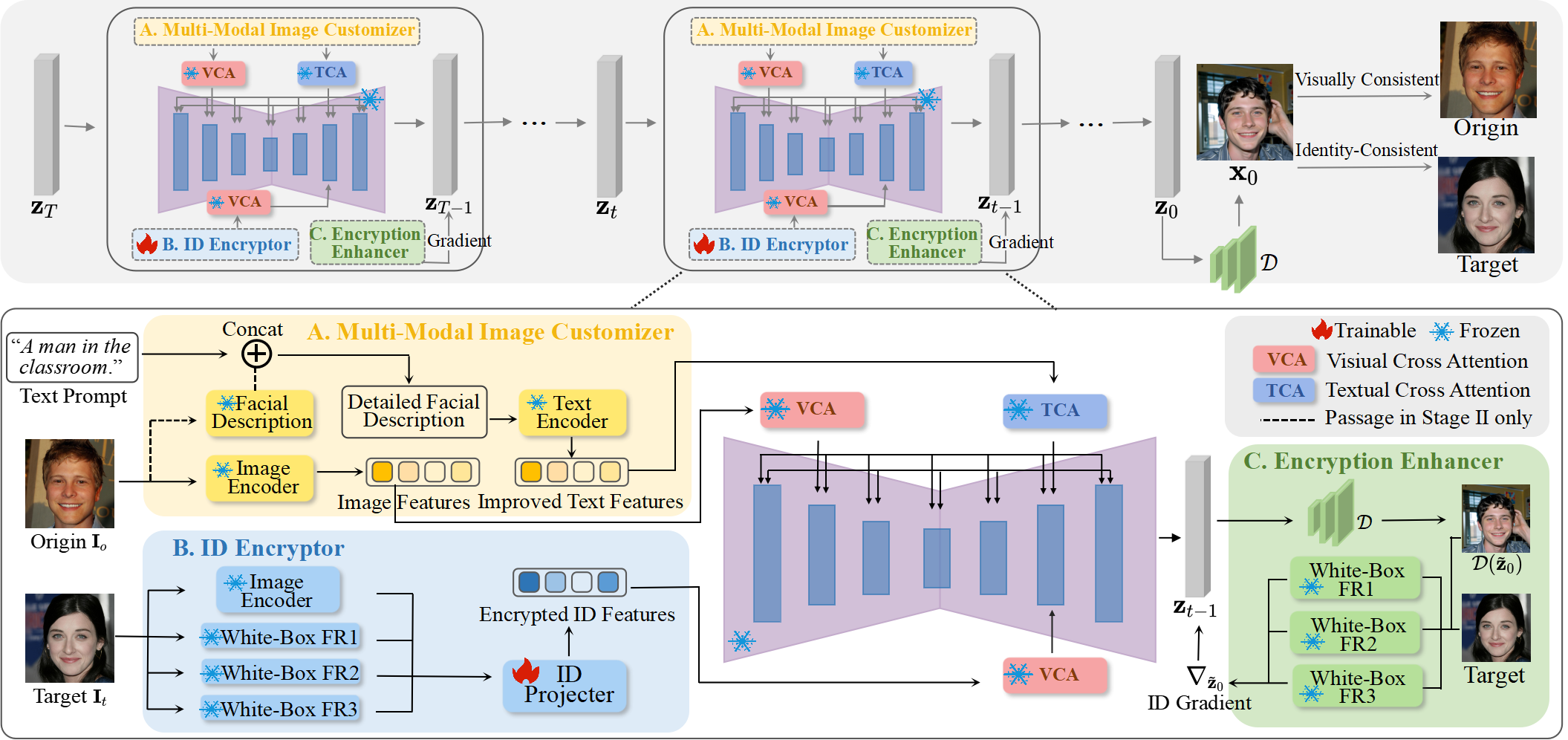}
   \vspace{-0.6cm}
   \caption{The overall framework of the proposed Adv-CPG. The framework comprises three key modules: [A] a multi-modal image customizer (MMIC), [B] an ID encryptor (En1), and [C] an encryption enhancer (En2). En1 and En2 enable incremental facial privacy protection, and MMIC achieves fine-grained and personalized portrait generation.}
   \vspace{-0.5cm}
   \label{fig:frame}
\end{figure*}

To address these two issues, this paper proposes a \textbf{C}ustomized \textbf{P}ortrait \textbf{G}eneration framework with facial \textbf{Adv}ersarial attacks (Adv-CPG). It generates portraits that can not only be abused by malicious FR systems, but also have fine-grained and customized features. Specifically, Adv-CPG aims to generate customized portraits with attack semantics (adversarial examples) that can deceive FR systems, leading to misidentifications and thus enabling facial privacy protection. As shown in Fig.~\ref{fig:frame}, the Multi-Modal Image Customizer (MMIC) is essential for fine-grained portrait generation, while the ID Encryptor (En1) and Encryption Enhancer (En2) offers progressive two-layer protection. The following are the key contributions:
\begin{itemize}
    \item{\textbf{First Customized Portrait Generation Framework with Facial Adversarial Attacks:} Adv-CPG is the first work introducing facial adversarial attacks into SD-based CPG. As illustrated in Fig.~\ref{fig:head}, the adversarial examples of Adv-CPG exhibit physical attack capabilities and stealthiness (for facial privacy protection), together with high fidelity and fine-grained features  (for portrait generation). }
    \item{\textbf{Progressive Facial Privacy Protection:} We accomplish progressive two-layer protection superior to any single-layer protection. Specifically, En1 elegantly realizes global protection by injecting target identities directly into the identity-dependent layer, and En2 achieves local protection by cost-effectively adding identity guidance.}
    \item{\textbf{Fine-Grained Customized Portrait Generation:} Adv-CPG achieves fine-grained and high-fidelity portrait generation, which is not available for other protection methods. This is attributed to the designed SD-based MMIC capable of generating controlled fine-grained facial features. Additionally, it allows for scenario and action customization with text prompts.}
\end{itemize}
Comparative experiments with state-of-the-art attack methods highlight the superior performance of Adv-CPG. Moreover, Adv-CPG facilitates tuning-free operation during testing, ensuring high performance at minimal training costs.


\section{Related Work}
\label{sec:related}
\textbf{Text-to-Image Diffusion Models.}
Advancement in diffusion models \cite{song2020denoising, rombach2022high, podell2023sdxl} has significantly propelled recent progress in text-conditional image and video generation \cite{guo2024pulid,peng2024portraitbooth,kong2024anymaker,gal2022image}. Current research is dedicated to improving the performance of these models in multiple ways, such as exploring high-quality large-scale text image datasets \cite{schuhmann2022laion,schuhmann2021laion,zhang2024data}, upgrading the base model~\cite{rombach2022high}, and improving the controllability of the model~\cite{ruiz2024hyperdreambooth,huang2024enhance,zhang2023variational,10684794}. Lately, the Stable Diffusion XL (SDXL)~\cite{podell2023sdxl} has been widely pursued due to its relatively low computational cost and impressive capability of portrait generation. Thus, Adv-CPG builds on SDXL but can be extended to other T2I diffusion models.

\textbf{Customized Portrait Generation.}
CPG \cite{li2024photomaker,wang2024instantid} aims to generate a portrait image that is visually consistent with the reference facial image while adhering to additional prompts. Depending on whether or not fine-tuning is performed during testing, mainstream CPG methods can be divided into two categories. Early approaches \cite{gal2022image, hulora,ruiz2024hyperdreambooth} require resource-intensive backpropagation, with each fine-tuning taking tens of minutes or hours to complete. Recent test-time tuning-free methods \cite{li2024photomaker,wang2024instantid,xiao2024fastcomposer,ye2023ip,he2024uniportrait,wang2024instantstyle_plus,huang2024consistentid} utilize extra modules to insert image semantics, giving pre-trained models visual inference comprehension, and are promising for large-scale deployment. Inspired by these, Adv-CPG follows the test-time tuning-free paradigm.

\textbf{Adversarial Attacks on Face Recognition.} The adversarial attack methods are generally categorized into noise-based methods and unrestricted methods. Conventional noise-based approaches \cite{goodfellow2014explaining, dong2018boosting, madry2017towards, yang2021towards} exploit adding $\ell_{p}$ boundary perturbations directly in the pixel space, which struggles with black-box transferability~\cite{xiao2021improving} and stealthiness~\cite{qiu2020semanticadv}. Recent unrestricted methods concentrate on learning perturbations in the semantic space, further improving transferability and stealthiness \cite{sharif2016accessorize, jia2022adv}. Based on adding extra makeup or not, mainstream unconstrained methods are further divided into two categories. Among them, makeup-based approaches \cite{komkov2021advhat, yin2021adv,hu2022protecting,shamshad2023clip2protect,li2024transferable, shamshad2024makeup,sun2024diffam} organize perturbations as makeup (or hat) for the adversarial examples, while facial semantic invariant methods \cite{liu2023diffprotect, zhang2024double, an2024sd4privacy, chow2025personalized} generate adversarial examples visually consistent with the original face. For clarity, Table~\ref{tab:1} shows the features of various methods (noised-based vs. unrestricted).

\section{Methodology}
\label{sec:method}
\textbf{Problem Definition.} A high-quality adversarial example should ensure that a human regards it as the same person as the original face (stealthiness), and meanwhile lead the malicious FR systems to misclassify it as another identity in black-box attack scenarios (physical attack capability and transferability). It involves a two-fold implication. Firstly, the original face $\mathbf{I}_o$ and the generated adversarial example $\mathbf{I}_{adv}$ should be visually consistent to human observers ($\mathbf{I}_{adv}= G(\mathbf{I}_o)$, $H(\mathbf{I}_{adv})\approx H(\mathbf{I}_o)$). Among it, $H(\cdot)$ denotes visual results by human observers, and $G(\cdot)$ indicates the strong generative model to be implemented by the proposed method, which generates the adversarial example $\mathbf{I}_{adv}$ conditional on the original face $\mathbf{I}_o$. Secondly, the target image $\mathbf{I}_t$ and the adversarial example $\mathbf{I}_{adv}$ should maintain identity consistent, which is formulated as follows
\begin{equation}
    \mathop{\arg}\mathop{\max}\limits_{\mathbf{I}_{adv}}F(\mathbf{I}_t, \mathbf{I}_{adv}),
\end{equation}
where $F(\cdot)$ denotes the identity similarity between this pair of faces measured by the face verification system. Generally, a FR system predicts that two faces belong to the same identity if $F(\cdot) \geq \tau$, where $\tau$ is the system threshold. In this paper, beyond the physical attack capability, transferability, and stealthiness, the proposed Adv-CPG also satisfies the personalized needs of users, which makes an effort to explore high-fidelity, refined facial generation, and natural background introduction for adversarial examples.

\subsection{Preliminaries}
\textbf{Stable Diffusion.} Stable Diffusion (SD) \cite{rombach2022high,zhang2023adding} has recently emerged as an increasingly popular T2I model. SD consists of four key components: an encoder $\mathcal{E}$ and a decoder $\mathcal{D}$ for mapping between the original pixel space and the low-dimensional latent space, a U-Net denoiser $\epsilon_{\theta}$ for generating denoised diffusion images, and a CLIP-based ~\cite{radford2021learning} pre-trained text encoder $\mathcal{T}$. Following the mainstream denoising diffusion probabilistic model (DDPM)~\cite{ho2020denoising}, SD is formulated as a forward diffusion process and a $T$-steps backward denoising process. In the diffusion phase, random Gaussian noise $\bm{\epsilon}$ is gradually added to the initial latent $\mathbf{z}_0$ to yield the noisy latent $\mathbf{z}_t$ at different timesteps $t\in\left[1, T\right]$. Then denoising U-Net is trained by the loss
\begin{equation}
    \mathcal{L}_{\rm SD} = \mathbb{E}_{\mathbf{z}_0,\mathbf{C},\bm{\epsilon}\sim\mathcal{N} (0,\mathbf{I}),t}\left[\|\bm{\epsilon}-\epsilon_{\theta}(\mathbf{z}_t,t,\mathbf{C})\|\right],
\end{equation}
where $\mathbf{C}$ is the optional additional condition embedding. 



\textbf{Image Prompt Adapter (IP-Adapter).} Recent test-time tuning-free CPG methods \cite{guo2024pulid,wang2024instantid} utilize the IP-Adapter~\cite{ye2023ip} to accomplish visual information injection, drawing widespread attention. Specifically, IP-Adapter extracts features of the reference image and feeds them into an additional cross-attention layer, which can be formulated as
\begin{equation}
\left\{  \begin{array}{lr}  
    \mathbf{z}'={\rm Attention}(\mathbf{Q},\mathbf{K},\mathbf{V})+\lambda\cdot{\rm Attention}(\mathbf{Q},\mathbf{K}^i,\mathbf{V}^i), \\
     \mathbf{K}=\mathbf{C}_t\mathbf{W}_K,  \mathbf{V}=\mathbf{C}_t\mathbf{W}_V, \mathbf{K}^i=\mathbf{C}_i\mathbf{W}_K^i,  \mathbf{V}^i=\mathbf{C}_i\mathbf{W}_V^i, 
    \end{array}  \right. 
\end{equation}
where ${\rm Attention}(\mathbf{Q},\mathbf{K},\mathbf{V})$ is the original textual cross-attention layer, ${\rm Attention}(\mathbf{Q},\mathbf{K}^i,\mathbf{V}^i)$ is the added visiual cross-attention layer, $\mathbf{Q}$ is the spatial feature, $\mathbf{C}_t$ and $\mathbf{C}_i$ are textual feature and image feature, and $\mathbf{W}_K$, $\mathbf{W}_V$, $\mathbf{W}_K^i$, and $\mathbf{W}_V^i$ are the trainable weights of the linear layers.

\begin{figure*}[t]
  \centering
   \includegraphics[width=1.0\linewidth]{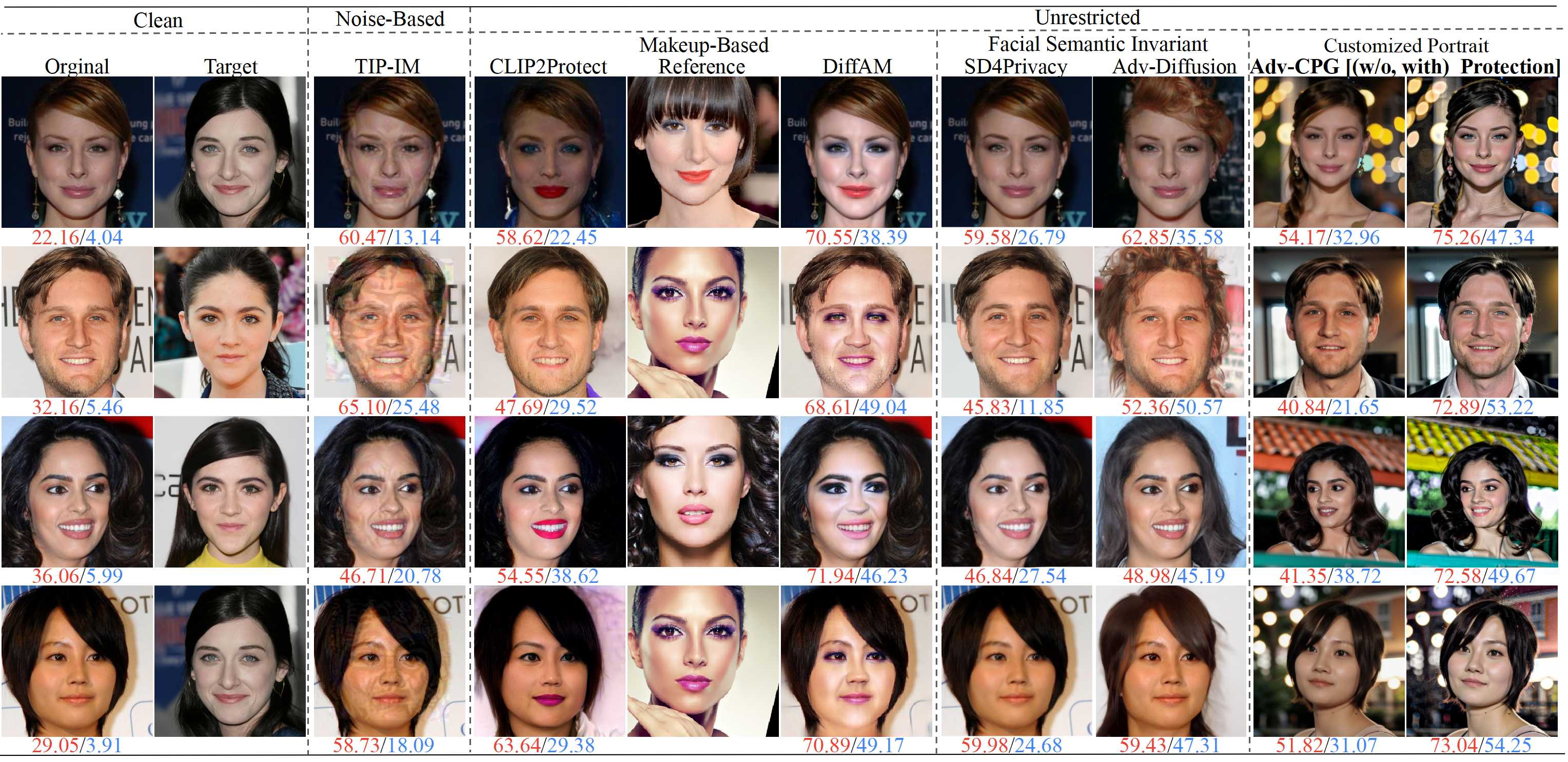}
   \vspace{-0.6cm}
   \caption{Visual quality comparison of adversarial examples generated by 4 types of adversarial methods on the CelebA-HQ dataset. The images generated by Adv-CPG implement both portrait customization and effective facial privacy protection. Red/blue (\textcolor[RGB]{234, 67, 35}{1}/\textcolor[RGB]{66, 133, 244}{2}) numbers below each image: confidence scores returned by Face++ and Aliyun (higher is better). Reference column: makeup reference for DiffAM. }
   \label{fig:Qualitative}
   \vspace{-0.3cm}
\end{figure*}

\subsection{Model Overview}
\textbf{Pipeline Overview.} To enable facial privacy protection and CPG, Adv-CPG is designed with three key modules: an ID encryptor (En1), an encryption enhancer (En2), and a multi-modal image customizer (MMIC), as shown in Fig.~\ref{fig:frame}. Furthermore, based on the working state of MMIC, the generation process of Adv-CPG can be divided into two stages: \textbf{1}) \textbf{Progressive Facial Privacy Protection Stage}: The original text prompt without enhancement is utilized for initial background introductions. In particular, En1 and En2 play important roles in achieving incremental facial privacy protection in this stage, which is stated in Section~\ref{subsec:Identity Learning}. \textbf{2}) \textbf{Fine-Grained Customized Portrait Generation Stage}: In this stage, the detailed facial feature generated by the MMIC is utilized to accomplish fine-grained portrait generation, while the other modules are integrated to maintain a balance between identity features and visual semantics. MMIC is explained in detail in Section~\ref{sub:Fine-Grained Face}.

\textbf{Training Phase.} Adv-CPG is built on the pre-trained SDXL~\cite{podell2023sdxl}, which is a SOTA T2I diffusion model. During training, only MMIC and En1 participate. Only the parameters of the ID projector in En1 will be optimized, while other parameters including the parameters of the SD model will be frozen. The loss during training is formulated as 
\begin{equation}
    \mathcal{L}_{\rm CPG} = \mathbb{E}_{\mathbf{z}_0,\mathbf{C}_{id},\mathbf{C}_i,\mathbf{C}_t,\bm{\epsilon},t}\left[\|\bm{\epsilon}-\epsilon_{\theta}(\mathbf{z}_t,t,\mathbf{C}_{id},\mathbf{C}_i,\mathbf{C}_t)\|\right],
\end{equation}
where $\mathbf{C}_{id}$ is the encrypted ID feature generated from En1, and $\mathbf{C}_i$ and $\mathbf{C}_t$ are the image feature and the improved text feature generated from MMIC, respectively. The augmented text feature $\mathbf{C}_t$ is formed by concatenating the original text prompt with the fine-grained facial description, which is described in Section~\ref{sub:Fine-Grained Face}. 

\textbf{Inference Phase.} During the inference phase, MMIC, En1, and En2 are involved. Meanwhile, the straightforward delayed primacy condition similar to Fastcomposer~\cite{xiao2024fastcomposer} and ConsistentID~\cite{huang2024consistentid} is employed. By replacing the initial text feature with the augmented text feature after a specific step, Adv-CPG effectively balances fine-grained facial preservation and personalized scene introduction. Notably, the generation process is divided into two stages depending on whether or not the augmented text is used, which corresponds to the dual-stage pipeline (\textit{1}) \textit{Progressive Facial Privacy Protection Stage} and \textit{2}) \textit{Fine-Grained Customized Portrait Generation Stage}).

\subsection{Progressive Facial Privacy Protection}
\label{subsec:Identity Learning}
In this stage, MMIC extracts the textual feature from the initial prompt and the visual feature from the original face. This stage does not use fine-grained facial descriptions, intending to enable the En1 and En2 to achieve facial privacy protection by learning the target identity. Specifically, the structures of these two modules are presented as follows. 

\textbf{ID Encryptor (En1).} The designed En1 is based on the IP-Adapter, which injects the target ID features directly into the SD model. Previous CPG researches \cite{li2024photomaker} merely require ensuring that the generated portraits are highly similar in visual semantics to the original face image. In contrast, Adv-CPG guarantees not only a high visual semantic similarity with the original face but also a high similarity of identity features between the generated adversarial example and the target image. In this study, we leverage both the fuzzy semantic information extracted by the CLIP image encoder and ID embeddings in the target images extracted by the white-box FR model. This approach enables the learning of a robust powerful ID feature to guide the learning process effectively.
Specifically, the powerful target ID feature generated by the Adv-CPG can be represented as
\begin{equation}
    \mathbf{C}_{id} = \alpha \cdot \mathcal{I}(\mathbf{I}_t) + \beta\cdot \sum_{i}({\rm FR}_i(\mathbf{I}_t)),
    \label{eq:ID Encryptor}
\end{equation}
where $\mathcal{I}$ is the pre-trained CLIP image encoder and ${\rm FR}_i(\cdot)$ denotes the $i$-th pre-trained white-box FR model. Then, the target ID feature is directly injected into the identity-related cross-attention layer like PuLID~\cite{guo2024pulid} and InstantID~\cite{wang2024instantid}. 

\textbf{Encryption Enhancer (En2).} Inspired by Adv-Diffusion~\cite{liu2024adv} and FreeTuner~\cite{xu2024freetuner}, additional ID guidance is introduced as an encryption enhancement. Specifically, in each denoising step, the current noise reconstructed image $\mathcal{D}(\widetilde{\mathbf{z}}_0)$ and the target image $\mathbf{I}_t$ are used to generate an ID gradient as 
\begin{equation}
\left\{  \begin{array}{lr}  
    \mathbf{g}_t =\lambda_s\nabla_{\widetilde{\mathbf{z}}_0}F(\mathcal{D}(\widetilde{\mathbf{z}}_0), \mathbf{I}_{t}), \\
    \hat{\mathbf{z}}_{t-1} := \mathbf{z}_{t-1} + \bm{\sigma}_t\mathbf{g}_t
\end{array}  \right. 
\end{equation}
where $F(\cdot)$ is the ID similarity of the image pairs, $\lambda_s$ is the guidance strength, and the generated ID gradient $\mathbf{g}_t$ is used to update the prediction noise for identity enhancement.

\subsection{Fine-Grained Customized Portrait Generation}
\label{sub:Fine-Grained Face}
After certain steps in the protection phase, fine-grained portrait generation will be implemented. In this stage, the detailed facial description in MMIC is utilized to generate the detailed features. En1 and En2 continued to be integrated for a balance between CPG and facial privacy protection. Specifically, the structure of MMIC is stated as follows.

 
\textbf{Multi-Modal Image Customizer (MMIC).} Observing that the initial IP-Adapter is susceptible to inconsistent facial details, the text feature is refined into fine-grained face text features. Initially, a multi-modal large model, LLaVA1.5~\cite{liu2024improved}, is introduced as the facial description model to obtain detailed textual facial description by the prompt "Describe this person’s facial features, including face, ears, eyes, nose, and mouth". Then the detailed facial description and the original text prompt are concatenated and fed into the CLIP text encoder for generating the augmented text feature $\mathbf{C}_t$. Meanwhile, for a high visual semantic similarity with the original face, the CLIP image encoder is employed on the original face to extract the visual semantic features $\mathbf{C}_i$. Then the features $\mathbf{C}_t$ and $\mathbf{C}_i$ obtained by MMIC are injected in parallel into the decoupled cross-attention layer for fine-grained portrait generation.


\section{Experiments}
\label{sec:experiments}

\subsection{Experimental Setting}
\label{subsec: Experimental Setting}

\begin{table*}[!t]
    \caption{Attack success rate (ASR \%) of black-box attack under the face verification task. For each column, the other three models are used as surrogates to generate the adversarial examples. -: non-existent. MF: acronym for MobileFace. Bolded and labeled in light blue (\fcolorbox[RGB]{232,240,254}[RGB]{232,240,254}{\textbf{1}}): the best. Underlined ($\underline{2}$): the second. \textcolor[RGB]{200,36,35}{$\uparrow$}/$\downarrow$: performance comparison of Adv-CPG vs. the prior SOTA.}
    \centering
    \vspace{-0.12cm}
    \scalebox{0.785}{\begin{tabular}{c|c|cccc|cccc|c}
        \toprule
        \textbf{Method} & \textbf{Dataset} & \multicolumn{4}{c}{\textbf{FFHQ}} & \multicolumn{4}{c|}{\textbf{CelebA-HQ}} &\textbf{Average}\\
         &ASR\% on FR Model $\uparrow$ & IR152  & IRSE50  & FaceNet & MF & IR152  & IRSE50  & FaceNet & MF   & \\
        \midrule
        Clean &-— & 3.14     & 4.73  & 1.49    & 8.96   & 4.64  & 5.74    & 1.06 & 13.27 &5.38\\
        \midrule
        &FGSM\footnotesize{ (ICLR’15)}~\cite{goodfellow2014explaining} & 9.86     & 48.53 & 4.23    & 51.48  & 12.07     & 45.81  & 1.35    & 53.02  &28.29\\
         &MI-FGSM\footnotesize{ (CVPR’18)}~\cite{dong2018boosting} & 46.31    & 69.24  & 20.62    & 69.26  & 46.53     & 70.57  & 27.09   & 58.94  &51.07\\
         Noise- &PGD\footnotesize{ (ICLR’18)}~\cite{madry2017towards} & 31.64     & 75.38  & 18.59   & 63.12  & 41.87     & 63.23  & 19.62    & 57.34 &46.35 \\
         Based&TI-DIM\footnotesize{ (CVPR’19)}~\cite{dong2019evading}  &43.57  &65.89 & 14.72  &53.31 & 35.13   & 62.38  & 13.68   & 52.84   &42.69\\
         &TIP-IM\footnotesize{ (ICCV’21)}~\cite{yang2021towards} & 46.25 &67.38 &59.82 &52.03  &41.26  &57.29 &39.07  &49.56   &51.58\\
        \cmidrule{1-11}
         &Adv-Hat\footnotesize{ (ICPR’21)}~\cite{komkov2021advhat} &13.77 &15.36 &5.26 &9.83   &5.04      &16.88   & 4.91    & 12.64   &10.46\\
         &Adv-Makeup\footnotesize{ (IJCAI’21)~\cite{yin2021adv}} &10.03 &25.57 & 1.08 & 20.38   & 12.68     & 19.95  & 1.37    & 22.11    &14.15\\
          &AMT-GAN\footnotesize{ (CVPR’22)}~\cite{hu2022protecting} & 11.52  & 56.02  & 9.74  & 41.32  & 12.09     & 53.26  & 4.87   & 47.95 &29.60\\
         Makeup&Clip2Protect\footnotesize{ (CVPR’23)}~\cite{shamshad2023clip2protect} &52.12  &86.53 &45.01  &76.29   & 47.63  & 80.96 &42.57    & 73.64  &63.09\\
         -Based&GIFT\footnotesize{ (ACMMM’24)}~\cite{li2024transferable} &69.72   &87.64   &54.49  &82.93 &$\underline{73.84}$   &83.72  &56.48 &$\underline{86.37}$ &74.40\\
         &DFPP\footnotesize{ (arXiv’24)}~\cite{shamshad2024makeup} &54.25 &$\underline{90.63}$ & 52.13 & 80.09 &46.38 & 80.59 & 45.37  & 72.13    &69.20\\
          &DiffAM\footnotesize{ (CVPR’24)}~\cite{sun2024diffam} &67.24 &90.30 &\cellcolor[RGB]{232,240,254} \textbf{64.96} &$\underline{89.56}$ &65.09  & \cellcolor[RGB]{232,240,254} \textbf{89.66}    &$\underline{62.99}$   &84.51 &$\underline{76.79}$ \\
         \cmidrule{1-11}
           &DiffProtect\footnotesize{ (arXiv’23)}~\cite{liu2023diffprotect} &57.62 &60.14 & 49.38 & 67.52   & 58.64     & 79.34  & 24.69    & 75.91    &59.16\\
         Facial&DPG\footnotesize{ (arXiv’24)}~\cite{zhang2024double} &34.87 &76.82 & 36.57 & 69.03  & 42.89  &62.47  & 35.83   & 66.42    &53.11\\
         Semantic&SD4Privacy\footnotesize{ (ICME’24)}~\cite{an2024sd4privacy} &51.31 &79.94  &43.57   &71.55  & 66.89  & 79.96    & 53.49 & 74.58 &65.16\\
         Invariant&Adv-Diffusion\footnotesize{(AAAI’24)}~\cite{liu2024adv} & 50.93     & 81.76  & 30.84   & 67.52  & 52.84     & 81.67 & 34.95    & 70.78 &58.91\\
         &P3-Mask\footnotesize{ (ECCV’25)}~\cite{chow2025personalized} &$\underline{73.18}$ &85.35 & 57.92 & 70.69   & 73.47     & 83.40  & 60.24   & 69.64    &71.74\\
         \cmidrule{1-11}
         Portraits &Adv-CPG\footnotesize{ (Ours)} &\cellcolor[RGB]{231,239,250} \textbf{75.26}\footnotesize\textcolor[RGB]{200,36,35}{$\uparrow$\textbf{2.1}} &\cellcolor[RGB]{232,240,254} \textbf{91.03}\footnotesize\textcolor[RGB]{200,36,35}{$\uparrow$\textbf{0.4}}   &$\underline{63.84}$\footnotesize$\downarrow$1.1    &\cellcolor[RGB]{232,240,254} \textbf{89.94}\footnotesize\textcolor[RGB]{200,36,35}{$\uparrow$\textbf{0.4}}  &\cellcolor[RGB]{232,240,254} \textbf{76.96}\footnotesize\textcolor[RGB]{200,36,35}{$\uparrow$\textbf{3.1}}     & $\underline{88.72}$\footnotesize$\downarrow$0.9 &\cellcolor[RGB]{232,240,254} \textbf{63.50}\footnotesize\textcolor[RGB]{200,36,35}{$\uparrow$\textbf{0.5}}    &\cellcolor[RGB]{232,240,254} \textbf{87.95}\footnotesize\textcolor[RGB]{200,36,35}{$\uparrow$\textbf{1.6}} &\cellcolor[RGB]{232,240,254} \textbf{79.65}\footnotesize\textcolor[RGB]{200,36,35}{$\uparrow$\textbf{2.9}}\\
        \bottomrule
    \end{tabular}}
    \label{table:total_res}
    \vspace{-0.15cm}
\end{table*}
\begin{table*}[!t]
    \caption{Attack success rate (ASR \%) of black-box attack under the face identification task. For each column, the other three models are used as surrogates to generate the adversarial examples. R1-T: Rank-1-Targeted. R5-T: Rank-5-Targeted. Bolded and labeled in light blue (\fcolorbox[RGB]{232,240,254}[RGB]{232,240,254}{\textbf{1}}): the best. Underlined ($\underline{2}$): the second. \textcolor[RGB]{200,36,35}{$\uparrow$}/$\downarrow$: performance comparison of Adv-CPG vs. the prior SOTA.}
    \vspace{-0.2cm}
    \centering
    \scalebox{0.785}{\begin{tabular}{c|c|cc|cc|cc|cc|cc}
        \toprule
        \textbf{Method} & \textbf{FR Model} & \multicolumn{2}{c}{\textbf{IR152}} & \multicolumn{2}{c}{\textbf{IRSE50}} & \multicolumn{2}{c}{\textbf{FaceNet}} & \multicolumn{2}{c|}{\textbf{MobileFace}} & \multicolumn{2}{c}{\textbf{Average}}\\
         &ASR\% $\uparrow$ & R1-T  & R5-T  & R1-T  & R5-T & R1-T  & R5-T  & R1-T  & R5-T  & R1-T  & R5-T  \\
        \midrule
        \multirow{3}*{Noise-based} &MI-FGSM\footnotesize{ (CVPR’18)} & 3.0     & 13.2  & 3.8   & 12.6  & 8.4     & 22.0  & 7.8    & 20.6  &5.75     &17.10 \\
         &TI-DIM\footnotesize{ (CVPR’19)} & 8.2     & 14.2  & 4.2    & 13.8  & 17.8     & 32.2  & 21.4    & 43.8 &12.90    & 26.00\\
         &TIP-IM\footnotesize{ (ICCV’21)} & 9.8     & 29.8  & 7.4    & 27.8  & 25.8     & 55.8  & 31.8   & 52.6 & 18.70    & 41.50\\
        \cmidrule{1-12}
        \multirow{5}*{Unrestricted} &Clip2Protect\footnotesize{ (CVPR’23)} & 17.8    & 51.8  & 10.6    & 35.2 & 27.2  & 54.4 & 39.0   & $\underline{67.8}$ & 23.65 & 52.30\\
         &GIFT\footnotesize{ (ACMMM’24)} &$\underline{21.2}$  &  
         \cellcolor[RGB]{232,240,254} \textbf{57.2}  & \cellcolor[RGB]{232,240,254}\textbf{34.6} & $\underline{49.4}$   & 33.2  & $\underline{65.6}$  & $\underline{41.2}$   & 67.6 & $\underline{32.55}$  &$\underline{59.95}$ \\
          &DFPP\footnotesize{ (arXiv’24)} & 14.8   &41.4  &13.6 &37.2  & 25.2  & 53.8  & 36.6   & 60.4 &22.55   &48.20 \\
         \cmidrule{2-12}
         &SD4Privacy\footnotesize{ (ICME’24)} & 15.6 & 26.8 &23.4 &41.2  &$\underline{33.6}$  & 53.8  &31.8  & 49.8 &26.10    &42.90 \\
         \cmidrule{2-12}
         &Adv-CPG\footnotesize{ (Ours)} & \cellcolor[RGB]{232,240,254}\textbf{24.4}\footnotesize\textcolor[RGB]{200,36,35}{$\uparrow$\textbf{3.2}}     &$\underline{56.4}$\footnotesize$\downarrow$0.8  &$\underline{33.8}$\footnotesize$\downarrow$0.8  &\cellcolor[RGB]{232,240,254}\textbf{51.2}\footnotesize\textcolor[RGB]{200,36,35}{$\uparrow$\textbf{1.8}}  &\cellcolor[RGB]{232,240,254}\textbf{36.6}\footnotesize\textcolor[RGB]{200,36,35}{$\uparrow$\textbf{3.0}}     & \cellcolor[RGB]{232,240,254}\textbf{67.4}\footnotesize\textcolor[RGB]{200,36,35}{$\uparrow$\textbf{1.8}} & \cellcolor[RGB]{232,240,254}\textbf{43.4}\footnotesize\textcolor[RGB]{200,36,35}{$\uparrow$\textbf{2.2}}    & \cellcolor[RGB]{232,240,254}\textbf{70.4}\footnotesize\textcolor[RGB]{200,36,35}{$\uparrow$\textbf{2.6}} & \cellcolor[RGB]{232,240,254}\textbf{34.55}\footnotesize\textcolor[RGB]{200,36,35}{$\uparrow$\textbf{2.0}} &\cellcolor[RGB]{232,240,254}\textbf{61.35}\footnotesize\textcolor[RGB]{200,36,35}{$\uparrow$\textbf{1.4}} \\
        \bottomrule
    \end{tabular}}
    \vspace{-0.25cm}
    \label{table:total_res2}
\end{table*}

\textbf{Implementation Details.} Adam~\cite{kinga2015method} is set as the optimizer, the learning rate is set to $10^{-4}$, and the training is performed on 4 NVIDIA 3090 GPUs with a batch size of 8. In addition, a 50-step DDIM~\cite{song2020denoising} sampler is employed and the scale of the classifier guidance setting is set to 5. In the training process, we remove 50\% of the background information from the characters with a 25\% probability to minimize interference. At the same time, we replace the original text embedding with zero text embedding with 10\% probability. In the inference process, delayed topic conditioning \cite{karras2019style, xiao2024fastcomposer} is employed to resolve conflicts between text embeddings, semantic information embeddings, and identity embeddings.

\textbf{Datasets.} During the training phase, we employ the fine-grained face dataset FGID~\cite{huang2024consistentid}, from which 30,000 CelebA~\cite{liu2015deep} and 42,500 SFHQ~\cite{david_beniaguev_2022_SFHQ} annotations are selected. During the testing phase, three publicly available face datasets are used following a similar protocol \cite{jia2022adv, liu2024adv}, including FFHQ~\cite{karras2019style}, CelebA-HQ~\cite{karras2018progressive}, and LFW~\cite{huang2008labeled}. Among them, FFHQ and CelebA-HQ are used for face verification and LFW are used for face identification. For face verification, 1000 images with different identities are randomly selected as source images and randomly divided into 5 groups for FFHQ and CelebA-HQ datasets. Meanwhile, we select 5 images for each group. Consequently, each dataset forms 5 groups with 200 source-target image pairs in each group. For face identification, 500 identities are randomly selected from the LFW and one image of each identity is put into the gallery set and another image is put into the probe set. The attack setting is similar to GIFT~\cite{li2024transferable}.

\textbf{Benchmark.} Multiple adversarial attack benchmarking schemes are implemented for more fair performance comparisons, including 5 noise-based methods and 12 unconstrained methods. For noise-based one, the chosen methods consist of FGSM~\cite{goodfellow2014explaining}, MI-FGSM~\cite{dong2018boosting}, PGD~\cite{madry2017towards}, TI-DIM~\cite{dong2019evading}, and TIP-IM~\cite{yang2021towards}. The 12 unconstrained methods can be further categorized into 7 makeup-based methods:
Adv-Hat~\cite{komkov2021advhat}, Adv-Makeup~\cite{yin2021adv}, AMT-GAN~\cite{hu2022protecting}, Clip2Protect~\cite{shamshad2023clip2protect}, GIFT~\cite{li2024transferable}, DFPP~\cite{shamshad2024makeup}, and DiffAM~\cite{sun2024diffam}, and 5 facial semantic invariant methods:
 DiffProtect~\cite{liu2023diffprotect}, DPG~\cite{zhang2024double}, SD4Privacy~\cite{an2024sd4privacy}, Adv-Diffusion~\cite{liu2024adv}, and P3-Mask~\cite{chow2025personalized}. Among them, the single-objective variant of TIP-IM~\cite{dong2019evading} is utilized in this experiment for a fair comparison.


\textbf{Target FR Models.} Following Clip2Protect, 4 widely used deep FR models are selected to evaluate the attack performance of Adv-CPG and SOTA attack methods in black-box settings. In particular, these models are IR152~\cite{he2016deep}, IRSE50~\cite{hu2018squeeze}, FaceNet~\cite{schroff2015facenet}, and MobileFace~\cite{deng2019arcface}. At the same time, two commercial FR APIs (Face++\footnote[1]{https://www.faceplusplus.com/face-comparing/} and Aliyun\footnote[2]{https://vision.aliyun.com/experience/detail?\&tagName=facebody\newline \&children=CompareFace}) are selected in the experiment to evaluate the performance of different adversarial attack methods.
\begin{figure*}[t]
	\centering
	\subcaptionbox*{}{
		\label{res_head}
		\includegraphics[width=0.97\linewidth]{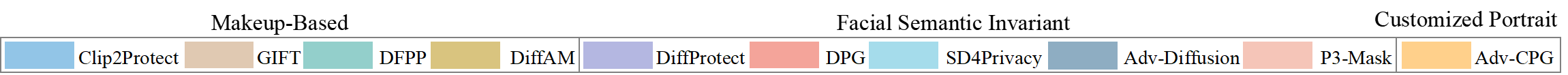}
	}\vspace{-0.3cm}
    \subcaptionbox*{Attack of CelebA-HQ on Face++}{
		\label{CelebA-HQ_on_Face}
		\includegraphics[width=0.2342\linewidth]{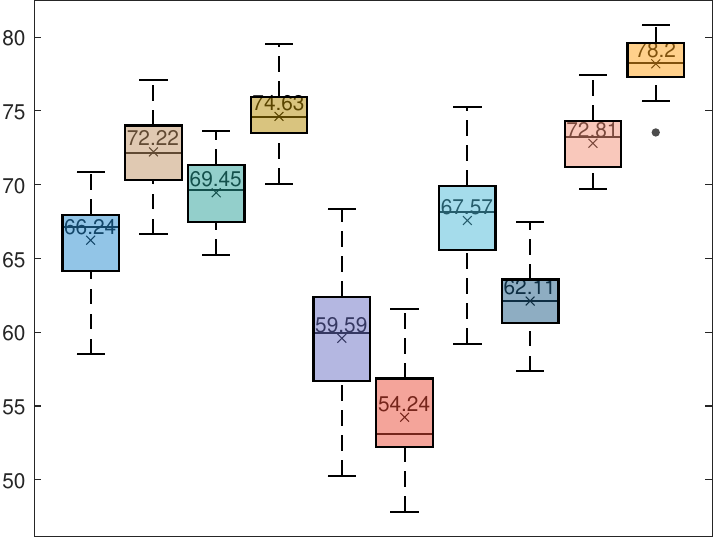}
	}\hfill	
	\subcaptionbox*{Attack of FFHQ on Face++}{
		\label{FFHQ_on_Face}
		\includegraphics[width=0.2342\linewidth]{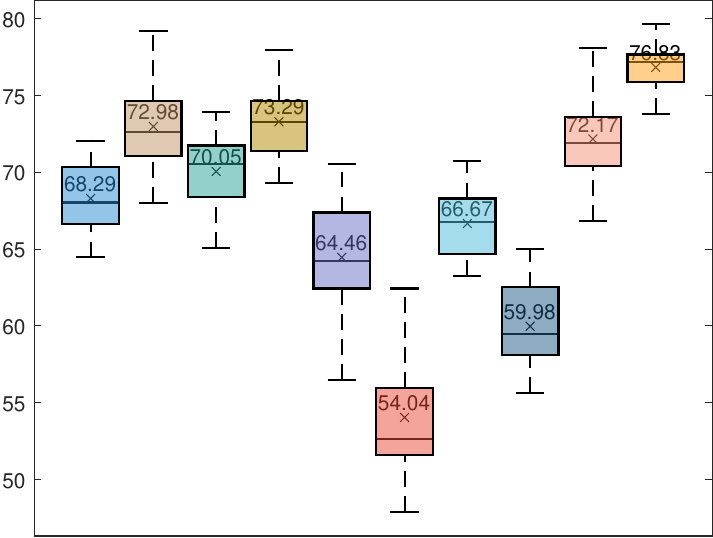}
	}\hfill
	\subcaptionbox*{Attack of CelebA-HQ on Aliyun}{
		\label{CelebA-HQ_on_Aliyun}
		\includegraphics[width=0.2342\linewidth]{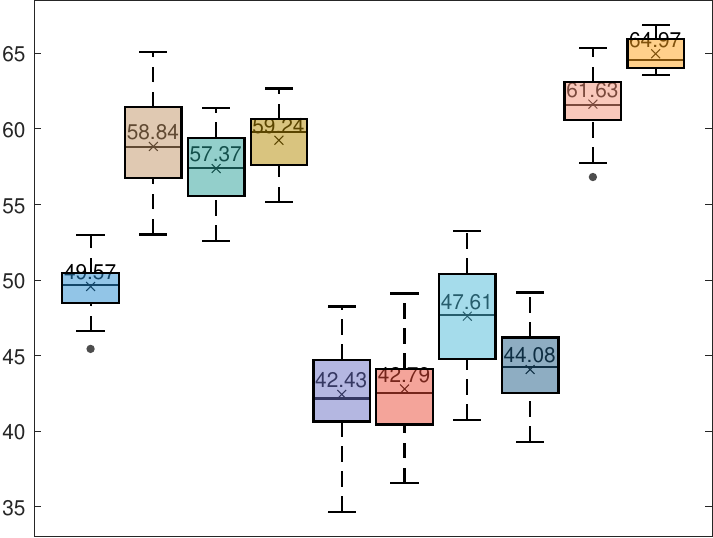}
	}\hfill
	\subcaptionbox*{Attack of FFHQ on Aliyun}{
		\label{FFHQ_on_Aliyun}
		\includegraphics[width=0.2342\linewidth]{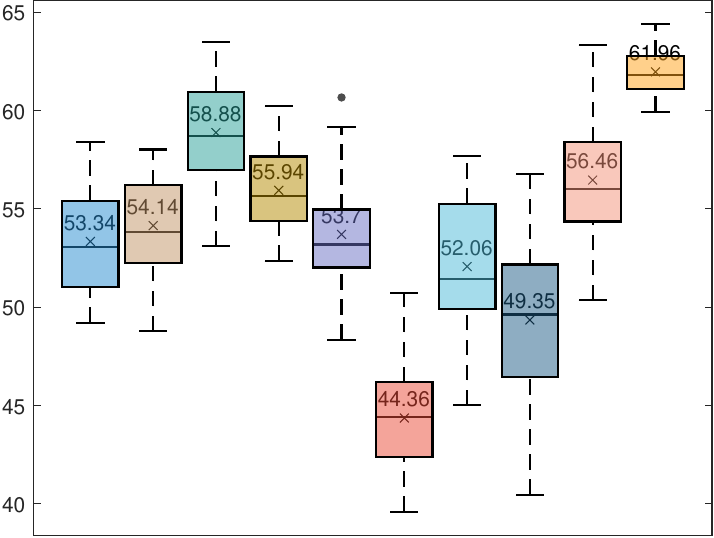}
	}\vspace{-0.2cm}
	\caption{The confidence scores ($\uparrow$) returned from commercial APIs, Face++ and Aliyun. Adv-CPG has higher and more stable confidence scores than state-of-the-art noise-based and makeup-based facial privacy protection methods.}
	\label{fig:commercial_APIs}
    \vspace{-0.05cm}
\end{figure*}

\begin{table*}[t]
    \vspace{-0.2cm}
    \caption{Quantitative evaluations of image quality for different adversarial examples. The FID, PSNR, and SSIM scores of black-box attacks on the CelebA-HQ dataset for 4 types of methods (noise-based, makeup-based, facial semantic invariant, and portrait-based).}
    \vspace{-0.15cm}
    \centering
    \scalebox{0.785}{\begin{tabular}{c|cc|cccc|cc|c}
        \toprule
       Method (Metric) &PGD &TIP-IM &Adv-Makeup &{AMT-GAN}
       &{CLIP2Protect} &{DiffAM} &{SD4Privacy} &{Adv-Diffusion} &\textbf{Adv-CPG} \\
        \cmidrule{1-10}
        FID $\downarrow$ &78.92 &38.7325 &4.2282 &34.5703 &26.1272 &26.1015 &26.4078 &22.5751 &26.0758\\
        PSNR $\uparrow$  &27.96 &33.2106 &34.5152 &19.5048 & 19.3542  & 20.5260 &26.8864 &28.8501 &29.9996\\
         SSIM$\uparrow$ &0.8573 &0.9236 &0.9850 &0.7924 & 0.6025 & 0.8861 &0.8092 &0.8046 &0.8978\\
        \bottomrule
    \end{tabular}}
    \vspace{-0.3cm}
    \label{table:image_quality}
\end{table*}


\textbf{Evaluation Metrics.} This work is evaluated from the privacy protection perspective (face verification and face identification) and image quality perspective. For face verification, the attack success rate (ASR) \cite{deb2020advfaces} (the ratio of successfully attacked adversarial examples to all adversarial examples) is reported as the evaluation metric. For face identification, the Rank-N target ASR is used as the metric, i.e., at least one of the top N images belongs to the target identity after ranking the similarity of all the images in the gallery set with the probe image. As for commercial FR APIs, the confidence scores returned by FR servers are directly used as the metric. To evaluate the quality of adversarial examples, the FID, PSNR, and SSIM scores are utilized as the quality metrics~\cite{heusel2017gans}. A lower FID, a larger PSNR, and a larger SSIM demonstrate a higher degree of similarity between the two compared images.


\subsection{Comparison Results}
\label{subsec: Comparison Results}

\textbf{Comparison on Black-Box Attacks.} For face verification experiments, the threshold $\tau$ is set as FAR@0.01 (false acceptance rate) for each selected FR model, i.e., IR152 (0.167), IRSE50 (0.241), FaceNet (0.409), and MobileFace (0.302). Table~\ref{table:total_res} reports the quantitative results of the proposed Adv-CPG and the 17 benchmark methods on the FFHQ and CelebA-HQ datasets against the dominant FR models under the black-box attack setting. In terms of ASR, the proposed method can achieve an average absolute gain of about 28.1\% and 2.86\% compared to the SOTA noise-based method (TIP-IM) and unconstrained method (DiffAM), respectively. In addition, most of the methods are weakly transferable on FaceNet due to the obvious differences in the FaceNet architecture. In contrast, Adv-CPG still generates robust adversarial examples with strong transferability, which demonstrates the effectiveness of facial privacy protection from the perspective of fine-grained face generation and customized background generation. For face identification experiments, Table~\ref{table:total_res2} reports the ASR results of the proposed Adv-CPG and 7 well-performing methods on the LFW dataset. Notably, since some works are initially trained to impersonate target identity in the face verification tasks, and others employ untargeted identity strategies, they are not included in Table~\ref{table:total_res2}. In most cases, the results of the proposed Adv-CPG are superior to recent approaches under Rank-1 and Rank-5 settings.


\textbf{Comparison on Image Quality.} Fig.~\ref{fig:Qualitative} illustrates the visual quality comparison of adversarial examples generated by different methods. We select 2-4 methods with superior performance from noised-based, unrestricted makeup-based, and unrestricted facial semantic invariant methods as the benchmark for comparison. In experiments, we prepare makeup text prompts (e.g., “red lipstick with blue eyeshadow”) to the CLIP2Protect method for generating the specified makeup, reference makeup images for the DiffAM method, and scene text prompts (e.g., “in the park”) to Adv-CPG for generating the specific scene. In contrast to the noise-based method TIP-IM, the adversarial examples of Adv-CPG are more natural, with no obvious noise patterns. In terms of unrestricted makeup-based approaches, CLIP2Protect generates reasonable and high-quality makeup, and DiffAM achieves outstanding performance in terms of precision and refinement. For the unconstrained facial semantic invariant approach, both SD4Privacy and Adv-Diffusion achieve very minimal changes to the facial part by operating at the semantic level. However, these methods only operate on the original face, which can only satisfy users' protection requirements but not more personalized demands. \textit{In contrast, Adv-CPG generates head-and-shoulders/half-body portraits with high fidelity of facial parts, high attack success rate, and natural overall appearance by a simple textual scene prompt}. As a result, this greatly satisfies the social and customization needs of users while successfully protecting facial privacy.

Meanwhile, Table~\ref{table:image_quality} quantitatively demonstrates the image quality evaluation results by reporting the FID, PSNR, and SSIM scores of different methods on the CelebA-HQ dataset. Among all quantitative evaluations, Adv-Makeup performs the best. The reason is that Adv-makeup only generates eye shadow, not full-face makeup, which makes minimal modifications to the image. However, as shown in Table~\ref{table:total_res}, the low ASR of Adv-Makeup in the experiment prevents it from standing out among diverse attack methods. In summary, Adv-CPG obtains comparatively low FID scores and high PSNR and SSIM scores, indicating that the images generated by Adv-CPG are relatively natural. 

\begin{figure}[t]
  \centering
   \includegraphics[width=1.02\linewidth]{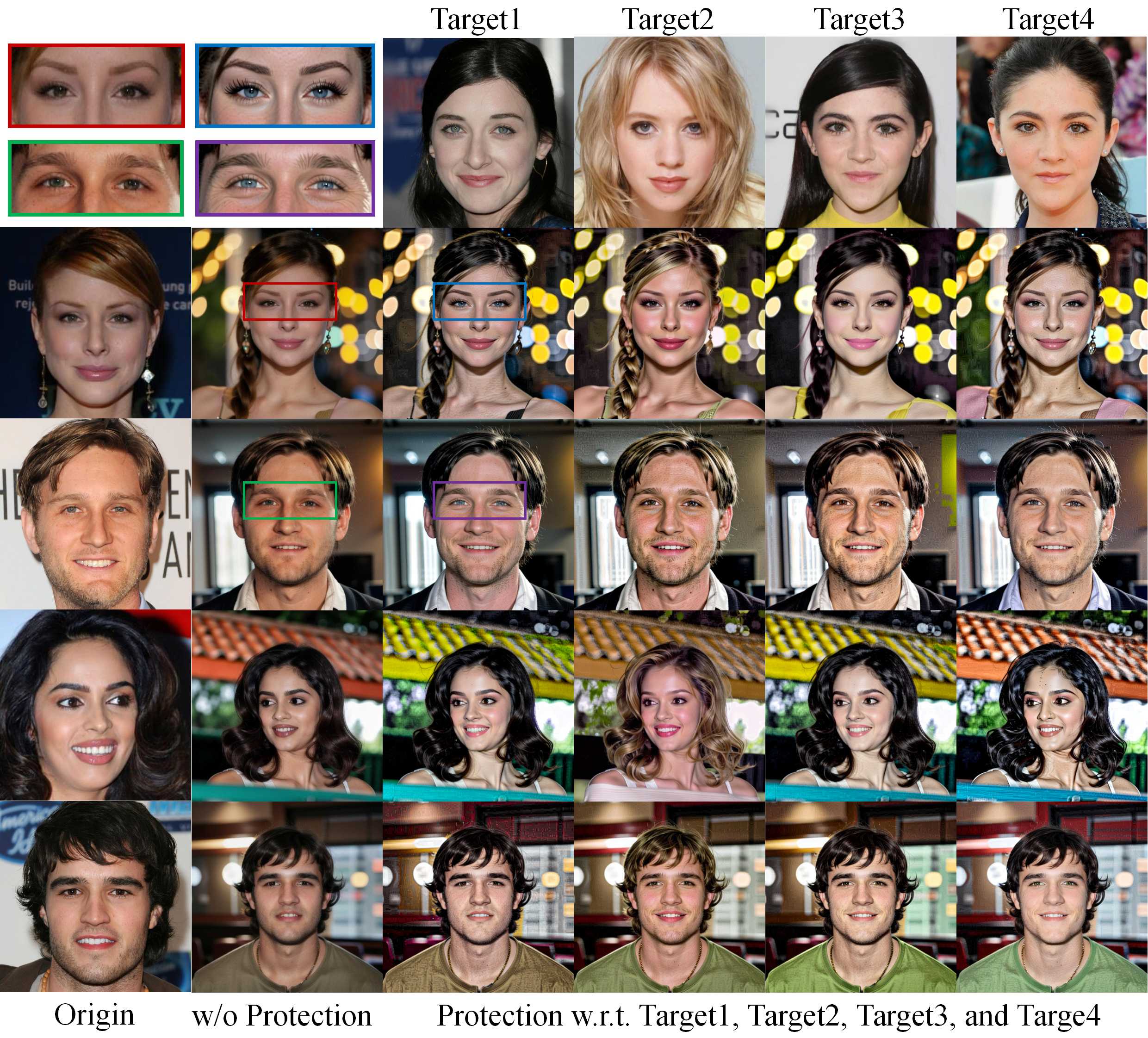}
   \vspace{-0.7cm}
   \caption{Visual comparison of adversarial examples for different target identities. First column: original face. Second column: personalized portrait without protection. Third/Fourth/Fifth/Sixth column: personalized portrait with protection to target-1/2/3/4. }
   \vspace{-0.4cm}
   \label{fig:aba1}
\end{figure}

\subsection{Attack Performance on Commercial APIs}
\label{subsec: Commercial APIs}
The quantitative attack results of different attack methods in face verification settings on commercial APIs (Face++ and Aliyun) are presented in Fig.~\ref{fig:commercial_APIs}. For benchmarking, recent methods are chosen, including 4 unrestricted makeup-based methods and 5 unrestricted facial semantical invariant methods. In experiments, we perform facial privacy protection on 100 randomly selected images from each of the CelebA-HQ and FFHQ datasets and report the confidence scores from the two APIs. The confidence score ranges from 0 to 100, with a higher score indicating a higher similarity between the adversarial example and the target image. The results demonstrate that the average confidence scores of Adv-CPG on each API are about 77.5 and 63.0, respectively, which is superior to current SOTA methods, and the attack effectiveness is comparatively stable. In addition, since these two APIs can effectively simulate facial privacy protection scenarios in the real world, it further evidences the high black-box attack capability of Adv-CPG.

\subsection{Ablation Study}
\label{subsec: Ablation Study}
\textbf{Robustness on Different Target Identities.} Generating adversarial examples with any given target face holds higher practical value. Therefore, four target images are randomly selected from the FFHQ and CelebA-HQ datasets 
 to evaluate the impact of target identity on the quality of the adversarial examples. As shown in Fig.~\ref{fig:aba1}, Adv-CPG generates high-fidelity attack examples for different targets, which indicates that Adv-CPG is robust to changes in target identity.

\begin{figure}[t]
  \centering
   \includegraphics[width=1.0\linewidth]{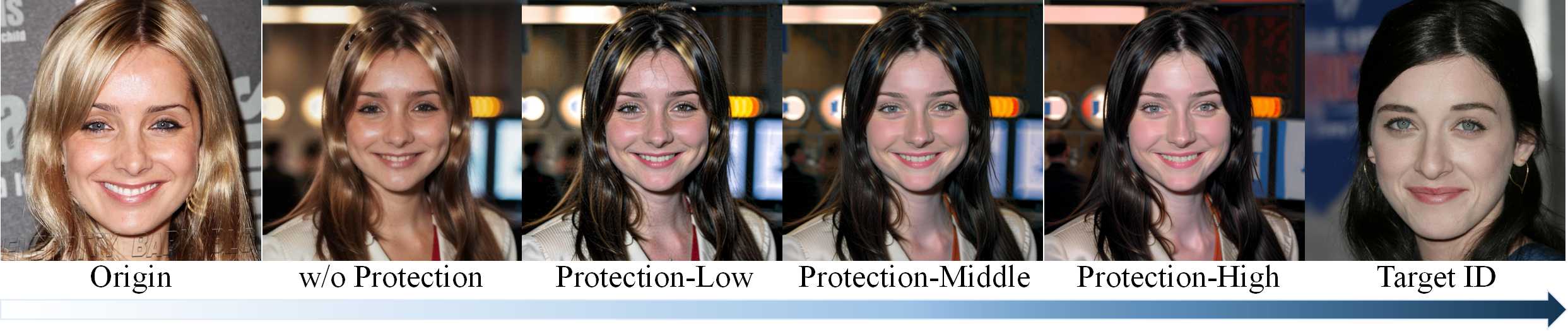}
   \vspace{-0.7cm}
   \caption{Qualitative impact of the encryption degree on the generated adversarial examples is demonstrated. As the encryption increases, the face features change towards the target identity.}
   \label{fig:aba2}
   \vspace{-0.3cm}
\end{figure}
\begin{table}[t]
    \caption{The impact of ID encryptor and encryptor enhancer on the ASR (\%$\uparrow$) of the generated adversarial examples compared to the CelebA-HQ dataset. MF: acronym for MobileFace.}
    \vspace{-0.1cm}
    \centering
    \scalebox{0.785}{\begin{tabular}{c|ccccc}
        \toprule
         Method (ASR) & IR152  & IRSE50  & FaceNet &MF & Average  \\
         \cmidrule{1-6}
         w/o Encryptor &54.67 &62.83  &39.62 & 67.04 &56.04  \\
        w/o Enhancer&69.35 & 80.74 &54.38 &79.59 &71.02  \\
        \cmidrule{1-6}
        \textbf{Adv-CPG}  &76.96 & 88.72 &63.50  &87.95 &79.28\\
        \bottomrule
    \end{tabular}}
    \vspace{-0.3cm}
    \label{table:abs3}
\end{table}

\textbf{Different levels of Identity Encryption.} Fig.~\ref{fig:aba2} illustrates the qualitative impact of the adversarial examples generated during a set of incremental identity encryptions (including ID Encryptor and encryption enhancer). As the level of encryption increases, the more similar the facial features in the target identity and the adversarial examples.

\textbf{Effectiveness of ID Encryptor and Enhancer.} To verify the effectiveness of the ID encryptor and encryption enhancer on the generated adversarial examples, three settings (no ID encryptor, no encryption enhancer, and the original setting) on the CelebA-HQ dataset are employed to evaluate the changes of the ASR. As shown in Table~\ref{table:abs3}, both the ID encryptor and encryption enhancer achieve facial privacy protection, especially the ID encryptor effectively achieves target learning by directly injecting the target identity.

\section{Conclusion}
\label{sec: Conclusion}
This paper presents Adv-CPG, a novel customized portrait generation framework with facial adversarial attacks. By injecting the target identity and adding additional identity guidance, Adv-CPG achieves progressive facial privacy protection. Further, by generating the detailed facial feature in the second stage, Adv-CPG achieves fine-grained portrait generation. Experimental results demonstrate that Adv-CPG surpasses current methods, enabling robust black-box attacks against diverse FR models and commercial APIs while achieving high-quality and customized portrait generation. Future work could explore the application of this facial privacy protection paradigm to other facial manipulation tasks, such as face editing and face swapping, to fully protect facial privacy in the age of artificial intelligence.

{
    \small
    \bibliographystyle{ieeenat_fullname}
    \bibliography{main}

\begin{thebibliography}{60}
\providecommand{\natexlab}[1]{#1}
\providecommand{\url}[1]{\texttt{#1}}
\expandafter\ifx\csname urlstyle\endcsname\relax
  \providecommand{\doi}[1]{doi: #1}\else
  \providecommand{\doi}{doi: \begingroup \urlstyle{rm}\Url}\fi

\bibitem[An et~al.(2024)An, Zhang, Wu, Lin, Gu, and Wang]{an2024sd4privacy}
Jinyang An, Wanqian Zhang, Dayan Wu, Zheng Lin, Jingzi Gu, and Weiping Wang.
\newblock Sd4privacy: Exploiting stable diffusion for protecting facial privacy.
\newblock In \emph{ICME}, pages 1--6, 2024.

\bibitem[Beniaguev(2022)]{david_beniaguev_2022_SFHQ}
David Beniaguev.
\newblock Synthetic faces high quality (sfhq) dataset, 2022.
\newblock https://doi.org/10.34740/kaggle/dsv/4737549, https://github.com/SelfishGene/SFHQ-dataset.

\bibitem[Chow et~al.(2025)Chow, Hu, Huang, and Liu]{chow2025personalized}
Ka-Ho Chow, Sihao Hu, Tiansheng Huang, and Ling Liu.
\newblock Personalized privacy protection mask against unauthorized facial recognition.
\newblock In \emph{ECCV}, pages 434--450, 2025.

\bibitem[Deb et~al.(2020)Deb, Zhang, and Jain]{deb2020advfaces}
Debayan Deb, Jianbang Zhang, and Anil~K Jain.
\newblock Advfaces: Adversarial face synthesis.
\newblock In \emph{IJCB}, pages 1--10, 2020.

\bibitem[Deng et~al.(2019)Deng, Guo, Xue, and Zafeiriou]{deng2019arcface}
Jiankang Deng, Jia Guo, Niannan Xue, and Stefanos Zafeiriou.
\newblock Arcface: Additive angular margin loss for deep face recognition.
\newblock In \emph{CVPR}, pages 4690--4699, 2019.

\bibitem[Dong et~al.(2018)Dong, Liao, Pang, Su, Zhu, Hu, and Li]{dong2018boosting}
Yinpeng Dong, Fangzhou Liao, Tianyu Pang, Hang Su, Jun Zhu, Xiaolin Hu, and Jianguo Li.
\newblock Boosting adversarial attacks with momentum.
\newblock In \emph{CVPR}, pages 9185--9193, 2018.

\bibitem[Dong et~al.(2019)Dong, Pang, Su, and Zhu]{dong2019evading}
Yinpeng Dong, Tianyu Pang, Hang Su, and Jun Zhu.
\newblock Evading defenses to transferable adversarial examples by translation-invariant attacks.
\newblock In \emph{CVPR}, pages 4312--4321, 2019.

\bibitem[Gal et~al.(2023)Gal, Alaluf, Atzmon, Patashnik, Bermano, Chechik, and Cohen-or]{gal2022image}
Rinon Gal, Yuval Alaluf, Yuval Atzmon, Or Patashnik, Amit~Haim Bermano, Gal Chechik, and Daniel Cohen-or.
\newblock An image is worth one word: Personalizing text-to-image generation using textual inversion.
\newblock In \emph{ICLR}, 2023.

\bibitem[Goodfellow et~al.(2015)Goodfellow, Shlens, and Szegedy]{goodfellow2014explaining}
Ian~J Goodfellow, Jonathon Shlens, and Christian Szegedy.
\newblock Explaining and harnessing adversarial examples.
\newblock In \emph{ICLR}, 2015.

\bibitem[Guo et~al.(2024)Guo, Wu, Chen, Chen, and He]{guo2024pulid}
Zinan Guo, Yanze Wu, Zhuowei Chen, Lang Chen, and Qian He.
\newblock Pulid: Pure and lightning id customization via contrastive alignment.
\newblock \emph{arXiv preprint arXiv:2404.16022}, 2024.

\bibitem[He et~al.(2024)He, Geng, and Bo]{he2024uniportrait}
Junjie He, Yifeng Geng, and Liefeng Bo.
\newblock Uniportrait: A unified framework for identity-preserving single-and multi-human image personalization.
\newblock \emph{arXiv preprint arXiv:2408.05939}, 2024.

\bibitem[He et~al.(2016)He, Zhang, Ren, and Sun]{he2016deep}
Kaiming He, Xiangyu Zhang, Shaoqing Ren, and Jian Sun.
\newblock Deep residual learning for image recognition.
\newblock In \emph{CVPR}, pages 770--778, 2016.

\bibitem[Heusel et~al.(2017)Heusel, Ramsauer, Unterthiner, Nessler, and Hochreiter]{heusel2017gans}
Martin Heusel, Hubert Ramsauer, Thomas Unterthiner, Bernhard Nessler, and Sepp Hochreiter.
\newblock Gans trained by a two time-scale update rule converge to a local nash equilibrium.
\newblock \emph{NeurIPS}, 30, 2017.

\bibitem[Ho et~al.(2020)Ho, Jain, and Abbeel]{ho2020denoising}
Jonathan Ho, Ajay Jain, and Pieter Abbeel.
\newblock Denoising diffusion probabilistic models.
\newblock \emph{NeurIPS}, 33:\penalty0 6840--6851, 2020.

\bibitem[Hu et~al.(2022{\natexlab{a}})Hu, Wallis, Allen-Zhu, Li, Wang, Wang, Chen, et~al.]{hulora}
Edward~J Hu, Phillip Wallis, Zeyuan Allen-Zhu, Yuanzhi Li, Shean Wang, Lu Wang, Weizhu Chen, et~al.
\newblock Lora: Low-rank adaptation of large language models.
\newblock In \emph{ICLR}, 2022{\natexlab{a}}.

\bibitem[Hu et~al.(2018)Hu, Shen, and Sun]{hu2018squeeze}
Jie Hu, Li Shen, and Gang Sun.
\newblock Squeeze-and-excitation networks.
\newblock In \emph{CVPR}, pages 7132--7141, 2018.

\bibitem[Hu et~al.(2022{\natexlab{b}})Hu, Liu, Zhang, Li, Zhang, Jin, and Wu]{hu2022protecting}
Shengshan Hu, Xiaogeng Liu, Yechao Zhang, Minghui Li, Leo~Yu Zhang, Hai Jin, and Libing Wu.
\newblock Protecting facial privacy: Generating adversarial identity masks via style-robust makeup transfer.
\newblock In \emph{CVPR}, pages 15014--15023, 2022{\natexlab{b}}.

\bibitem[Huang et~al.(2008)Huang, Mattar, Berg, and Learned-Miller]{huang2008labeled}
Gary~B Huang, Marwan Mattar, Tamara Berg, and Eric Learned-Miller.
\newblock Labeled faces in the wild: A database forstudying face recognition in unconstrained environments.
\newblock In \emph{Workshop on faces in'Real-Life'Images: detection, alignment, and recognition}, 2008.

\bibitem[Huang et~al.(2024{\natexlab{a}})Huang, Dong, Song, Li, Zhou, Cheng, Liao, Chen, Yan, Liao, et~al.]{huang2024consistentid}
Jiehui Huang, Xiao Dong, Wenhui Song, Hanhui Li, Jun Zhou, Yuhao Cheng, Shutao Liao, Long Chen, Yiqiang Yan, Shengcai Liao, et~al.
\newblock Consistentid: Portrait generation with multimodal fine-grained identity preserving.
\newblock \emph{arXiv preprint arXiv:2404.16771}, 2024{\natexlab{a}}.

\bibitem[Huang et~al.(2024{\natexlab{b}})Huang, Zhang, and Li]{huang2024enhance}
Sida Huang, Hongyuan Zhang, and Xuelong Li.
\newblock Enhance vision-language alignment with noise.
\newblock \emph{arXiv preprint arXiv:2412.10817}, 2024{\natexlab{b}}.

\bibitem[Jia et~al.(2022)Jia, Yin, Yao, Ding, Shen, Yang, and Ma]{jia2022adv}
Shuai Jia, Bangjie Yin, Taiping Yao, Shouhong Ding, Chunhua Shen, Xiaokang Yang, and Chao Ma.
\newblock Adv-attribute: Inconspicuous and transferable adversarial attack on face recognition.
\newblock \emph{NeurIPS}, 35:\penalty0 34136--34147, 2022.

\bibitem[Karras et~al.(2018)Karras, Aila, Laine, and Lehtinen]{karras2018progressive}
Tero Karras, Timo Aila, Samuli Laine, and Jaakko Lehtinen.
\newblock Progressive growing of {GAN}s for improved quality, stability, and variation.
\newblock In \emph{ICLR}, pages 1--12, 2018.

\bibitem[Karras et~al.(2019)Karras, Laine, and Aila]{karras2019style}
Tero Karras, Samuli Laine, and Timo Aila.
\newblock A style-based generator architecture for generative adversarial networks.
\newblock In \emph{CVPR}, pages 4401--4410, 2019.

\bibitem[Kinga et~al.(2015)Kinga, Adam, et~al.]{kinga2015method}
D Kinga, Jimmy~Ba Adam, et~al.
\newblock A method for stochastic optimization.
\newblock In \emph{ICLR}, page~6, 2015.

\bibitem[Komkov and Petiushko(2021)]{komkov2021advhat}
Stepan Komkov and Aleksandr Petiushko.
\newblock Advhat: Real-world adversarial attack on arcface face id system.
\newblock In \emph{ICPR}, pages 819--826, 2021.

\bibitem[Kong et~al.(2024)Kong, Wu, Hu, Han, Peng, Xu, Luo, Zhang, Wang, and Fu]{kong2024anymaker}
Lingjie Kong, Kai Wu, Xiaobin Hu, Wenhui Han, Jinlong Peng, Chengming Xu, Donghao Luo, Jiangning Zhang, Chengjie Wang, and Yanwei Fu.
\newblock Anymaker: Zero-shot general object customization via decoupled dual-level id injection.
\newblock \emph{arXiv preprint arXiv:2406.11643}, 2024.

\bibitem[Li et~al.(2024{\natexlab{a}})Li, Wang, Zhang, Zhou, Hu, and pei Xiaobing]{li2024transferable}
Minghui Li, Jiangxiong Wang, Hao Zhang, Ziqi Zhou, Shengshan Hu, and pei Xiaobing.
\newblock Transferable adversarial facial images for privacy protection.
\newblock In \emph{ACM MM}, 2024{\natexlab{a}}.

\bibitem[Li et~al.(2024{\natexlab{b}})Li, Cao, Wang, Qi, Cheng, and Shan]{li2024photomaker}
Zhen Li, Mingdeng Cao, Xintao Wang, Zhongang Qi, Ming-Ming Cheng, and Ying Shan.
\newblock Photomaker: Customizing realistic human photos via stacked id embedding.
\newblock In \emph{CVPR}, pages 8640--8650, 2024{\natexlab{b}}.

\bibitem[Liu et~al.(2024{\natexlab{a}})Liu, Wang, Peng, Wang, Hu, and Gao]{liu2024adv}
Decheng Liu, Xijun Wang, Chunlei Peng, Nannan Wang, Ruimin Hu, and Xinbo Gao.
\newblock Adv-diffusion: imperceptible adversarial face identity attack via latent diffusion model.
\newblock In \emph{AAAI}, pages 3585--3593, 2024{\natexlab{a}}.

\bibitem[Liu et~al.(2024{\natexlab{b}})Liu, Li, Li, and Lee]{liu2024improved}
Haotian Liu, Chunyuan Li, Yuheng Li, and Yong~Jae Lee.
\newblock Improved baselines with visual instruction tuning.
\newblock In \emph{CVPR}, pages 26296--26306, 2024{\natexlab{b}}.

\bibitem[Liu et~al.(2023)Liu, Lau, and Chellappa]{liu2023diffprotect}
Jiang Liu, Chun~Pong Lau, and Rama Chellappa.
\newblock Diffprotect: Generate adversarial examples with diffusion models for facial privacy protection.
\newblock \emph{arXiv preprint arXiv:2305.13625}, 2023.

\bibitem[Liu et~al.(2015)Liu, Luo, Wang, and Tang]{liu2015deep}
Ziwei Liu, Ping Luo, Xiaogang Wang, and Xiaoou Tang.
\newblock Deep learning face attributes in the wild.
\newblock In \emph{ICCV}, pages 3730--3738, 2015.

\bibitem[Madry et~al.(2018)Madry, Makelov, Schmidt, Tsipras, and Vladu]{madry2017towards}
Aleksander Madry, Aleksandar Makelov, Ludwig Schmidt, Dimitris Tsipras, and Adrian Vladu.
\newblock Towards deep learning models resistant to adversarial attacks.
\newblock In \emph{ICLR}, 2018.

\bibitem[Peng et~al.(2024)Peng, Zhu, Jiang, Tai, Luo, Zhang, Lin, Jin, Wang, and Ji]{peng2024portraitbooth}
Xu Peng, Junwei Zhu, Boyuan Jiang, Ying Tai, Donghao Luo, Jiangning Zhang, Wei Lin, Taisong Jin, Chengjie Wang, and Rongrong Ji.
\newblock Portraitbooth: A versatile portrait model for fast identity-preserved personalization.
\newblock In \emph{CVPR}, pages 27080--27090, 2024.

\bibitem[Podell et~al.(2024)Podell, English, Lacey, Blattmann, Dockhorn, M{\"u}ller, Penna, and Rombach]{podell2023sdxl}
Dustin Podell, Zion English, Kyle Lacey, Andreas Blattmann, Tim Dockhorn, Jonas M{\"u}ller, Joe Penna, and Robin Rombach.
\newblock {SDXL}: Improving latent diffusion models for high-resolution image synthesis.
\newblock In \emph{ICLR}, 2024.

\bibitem[Qiu et~al.(2020)Qiu, Xiao, Yang, Yan, Lee, and Li]{qiu2020semanticadv}
Haonan Qiu, Chaowei Xiao, Lei Yang, Xinchen Yan, Honglak Lee, and Bo Li.
\newblock Semanticadv: Generating adversarial examples via attribute-conditioned image editing.
\newblock In \emph{ECCV}, pages 19--37, 2020.

\bibitem[Radford et~al.(2021)Radford, Kim, Hallacy, Ramesh, Goh, Agarwal, Sastry, Askell, Mishkin, Clark, et~al.]{radford2021learning}
Alec Radford, Jong~Wook Kim, Chris Hallacy, Aditya Ramesh, Gabriel Goh, Sandhini Agarwal, Girish Sastry, Amanda Askell, Pamela Mishkin, Jack Clark, et~al.
\newblock Learning transferable visual models from natural language supervision.
\newblock In \emph{ICML}, pages 8748--8763, 2021.

\bibitem[Rombach et~al.(2022)Rombach, Blattmann, Lorenz, Esser, and Ommer]{rombach2022high}
Robin Rombach, Andreas Blattmann, Dominik Lorenz, Patrick Esser, and Bj{\"o}rn Ommer.
\newblock High-resolution image synthesis with latent diffusion models.
\newblock In \emph{CVPR}, pages 10684--10695, 2022.

\bibitem[Ruiz et~al.(2024)Ruiz, Li, Jampani, Wei, Hou, Pritch, Wadhwa, Rubinstein, and Aberman]{ruiz2024hyperdreambooth}
Nataniel Ruiz, Yuanzhen Li, Varun Jampani, Wei Wei, Tingbo Hou, Yael Pritch, Neal Wadhwa, Michael Rubinstein, and Kfir Aberman.
\newblock Hyperdreambooth: Hypernetworks for fast personalization of text-to-image models.
\newblock In \emph{CVPR}, pages 6527--6536, 2024.

\bibitem[Schroff et~al.(2015)Schroff, Kalenichenko, and Philbin]{schroff2015facenet}
Florian Schroff, Dmitry Kalenichenko, and James Philbin.
\newblock Facenet: A unified embedding for face recognition and clustering.
\newblock In \emph{CVPR}, pages 815--823, 2015.

\bibitem[Schuhmann et~al.(2021)Schuhmann, Vencu, Beaumont, Kaczmarczyk, Mullis, Katta, Coombes, Jitsev, and Komatsuzaki]{schuhmann2021laion}
Christoph Schuhmann, Richard Vencu, Romain Beaumont, Robert Kaczmarczyk, Clayton Mullis, Aarush Katta, Theo Coombes, Jenia Jitsev, and Aran Komatsuzaki.
\newblock Laion-400m: Open dataset of clip-filtered 400 million image-text pairs.
\newblock \emph{arXiv preprint arXiv:2111.02114}, 2021.

\bibitem[Schuhmann et~al.(2022)Schuhmann, Beaumont, Vencu, Gordon, Wightman, Cherti, Coombes, Katta, Mullis, Wortsman, et~al.]{schuhmann2022laion}
Christoph Schuhmann, Romain Beaumont, Richard Vencu, Cade Gordon, Ross Wightman, Mehdi Cherti, Theo Coombes, Aarush Katta, Clayton Mullis, Mitchell Wortsman, et~al.
\newblock Laion-5b: An open large-scale dataset for training next generation image-text models.
\newblock \emph{NeurIPS}, 35:\penalty0 25278--25294, 2022.

\bibitem[Shamshad et~al.(2023)Shamshad, Naseer, and Nandakumar]{shamshad2023clip2protect}
Fahad Shamshad, Muzammal Naseer, and Karthik Nandakumar.
\newblock Clip2protect: Protecting facial privacy using text-guided makeup via adversarial latent search.
\newblock In \emph{CVPR}, pages 20595--20605, 2023.

\bibitem[Shamshad et~al.(2024)Shamshad, Naseer, and Nandakumar]{shamshad2024makeup}
Fahad Shamshad, Muzammal Naseer, and Karthik Nandakumar.
\newblock Makeup-guided facial privacy protection via untrained neural network priors.
\newblock \emph{arXiv preprint arXiv:2408.12387}, 2024.

\bibitem[Sharif et~al.(2016)Sharif, Bhagavatula, Bauer, and Reiter]{sharif2016accessorize}
Mahmood Sharif, Sruti Bhagavatula, Lujo Bauer, and Michael~K Reiter.
\newblock Accessorize to a crime: Real and stealthy attacks on state-of-the-art face recognition.
\newblock In \emph{ACM CCS}, pages 1528--1540, 2016.

\bibitem[Song et~al.(2021)Song, Meng, and Ermon]{song2020denoising}
Jiaming Song, Chenlin Meng, and Stefano Ermon.
\newblock Denoising diffusion implicit models.
\newblock In \emph{ICLR}, 2021.

\bibitem[Sun et~al.(2024)Sun, Yu, Xie, Li, and Zhang]{sun2024diffam}
Yuhao Sun, Lingyun Yu, Hongtao Xie, Jiaming Li, and Yongdong Zhang.
\newblock Diffam: Diffusion-based adversarial makeup transfer for facial privacy protection.
\newblock In \emph{CVPR}, pages 24584--24594, 2024.

\bibitem[Wang et~al.(2024{\natexlab{a}})Wang, Xing, Huang, Ai, Wang, and Bai]{wang2024instantstyle_plus}
Haofan Wang, Peng Xing, Renyuan Huang, Hao Ai, Qixun Wang, and Xu Bai.
\newblock Instantstyle-plus: Style transfer with content-preserving in text-to-image generation.
\newblock \emph{arXiv preprint arXiv:2407.00788}, 2024{\natexlab{a}}.

\bibitem[Wang et~al.(2024{\natexlab{b}})Wang, Zhang, Wang, and Yuan]{10684794}
Junying Wang, Hongyuan Zhang, Hongwei Wang, and Yuan Yuan.
\newblock Graph convolutional network with self-augmented weights for semi-supervised multi-view learning.
\newblock \emph{IEEE TNNLS}, pages 1--14, 2024{\natexlab{b}}.

\bibitem[Wang et~al.(2024{\natexlab{c}})Wang, Bai, Wang, Qin, Chen, Li, Tang, and Hu]{wang2024instantid}
Qixun Wang, Xu Bai, Haofan Wang, Zekui Qin, Anthony Chen, Huaxia Li, Xu Tang, and Yao Hu.
\newblock Instantid: Zero-shot identity-preserving generation in seconds.
\newblock \emph{arXiv preprint arXiv:2401.07519}, 2024{\natexlab{c}}.

\bibitem[Xiao et~al.(2024)Xiao, Yin, Freeman, Durand, and Han]{xiao2024fastcomposer}
Guangxuan Xiao, Tianwei Yin, William~T Freeman, Fr{\'e}do Durand, and Song Han.
\newblock Fastcomposer: Tuning-free multi-subject image generation with localized attention.
\newblock \emph{IJCV}, pages 1--20, 2024.

\bibitem[Xiao et~al.(2021)Xiao, Gao, Fu, Dong, Gao, Zhang, Zhou, and Zhu]{xiao2021improving}
Zihao Xiao, Xianfeng Gao, Chilin Fu, Yinpeng Dong, Wei Gao, Xiaolu Zhang, Jun Zhou, and Jun Zhu.
\newblock Improving transferability of adversarial patches on face recognition with generative models.
\newblock In \emph{CVPR}, pages 11845--11854, 2021.

\bibitem[Xu et~al.(2024)Xu, Wang, Xiao, Liu, and Chen]{xu2024freetuner}
Youcan Xu, Zhen Wang, Jun Xiao, Wei Liu, and Long Chen.
\newblock Freetuner: Any subject in any style with training-free diffusion.
\newblock \emph{arXiv preprint arXiv:2405.14201}, 2024.

\bibitem[Yang et~al.(2021)Yang, Dong, Pang, Su, Zhu, Chen, and Xue]{yang2021towards}
Xiao Yang, Yinpeng Dong, Tianyu Pang, Hang Su, Jun Zhu, Yuefeng Chen, and Hui Xue.
\newblock Towards face encryption by generating adversarial identity masks.
\newblock In \emph{ICCV}, pages 3897--3907, 2021.

\bibitem[Ye et~al.(2023)Ye, Zhang, Liu, Han, and Yang]{ye2023ip}
Hu Ye, Jun Zhang, Sibo Liu, Xiao Han, and Wei Yang.
\newblock Ip-adapter: Text compatible image prompt adapter for text-to-image diffusion models.
\newblock \emph{arXiv preprint arXiv:2308.06721}, 2023.

\bibitem[Yin et~al.(2021)Yin, Wang, Yao, Guo, Kong, Ding, Li, and Liu]{yin2021adv}
Bangjie Yin, Wenxuan Wang, Taiping Yao, Junfeng Guo, Zelun Kong, Shouhong Ding, Jilin Li, and Cong Liu.
\newblock Adv-makeup: A new imperceptible and transferable attack on face recognition.
\newblock In \emph{IJCAI}, pages 1252--1258, 2021.

\bibitem[Zhang et~al.(2023{\natexlab{a}})Zhang, Huang, and Li]{zhang2023variational}
Hongyuan Zhang, Sida Huang, and Xuelong Li.
\newblock Variational positive-incentive noise: How noise benefits models.
\newblock \emph{arXiv preprint arXiv:2306.07651}, 2023{\natexlab{a}}.

\bibitem[Zhang et~al.(2024{\natexlab{a}})Zhang, Xu, Huang, and Li]{zhang2024data}
Hongyuan Zhang, Yanchen Xu, Sida Huang, and Xuelong Li.
\newblock Data augmentation of contrastive learning is estimating positive-incentive noise.
\newblock \emph{arXiv preprint arXiv:2408.09929}, 2024{\natexlab{a}}.

\bibitem[Zhang et~al.(2023{\natexlab{b}})Zhang, Rao, and Agrawala]{zhang2023adding}
Lvmin Zhang, Anyi Rao, and Maneesh Agrawala.
\newblock Adding conditional control to text-to-image diffusion models.
\newblock In \emph{ICCV}, pages 3836--3847, 2023{\natexlab{b}}.

\bibitem[Zhang et~al.(2024{\natexlab{b}})Zhang, Ye, Shen, Xie, Liu, Deng, and Tang]{zhang2024double}
Yunming Zhang, Dengpan Ye, Sipeng Shen, Caiyun Xie, Ziyi Liu, Jiacheng Deng, and Long Tang.
\newblock Double privacy guard: Robust traceable adversarial watermarking against face recognition.
\newblock \emph{arXiv preprint arXiv:2404.14693}, 2024{\natexlab{b}}.

\end{thebibliography}
}

\clearpage
\setcounter{page}{1}
\maketitlesupplementary

\section{Additional Algorithm Showcase}
\label{sec:Additional_Algorithm}
Based on whether or not the augmented text is used, the generation process of Adv-CPG is divided into two stages: \textbf{1}) \textbf{Progressive Facial Privacy Protection Stage}: In this stage, unenhanced raw text prompt is utilized for initial context introduction. Since the fine-grained facial description is not used, the designed facial privacy protection modules (ID encryptor and encryption enhancer) play an important role in progressive protection, and meanwhile the original textual prompt enables effective background generation. \textbf{2}) \textbf{Fine-Grained Customized Portrait Generation Stage}: In this phase, fine-grained face generation is achieved by utilizing detailed facial features and integrating other modules. This accomplishes not only a balance of visual semantics between the human body and the background, but also a balance between facial privacy protection and portrait generation.
The overall framework of the proposed Adv-CPG to perform facial privacy protection
and customized portrait generation is shown in Fig. \ref{fig:frame}, and the algorithm of the generation process is summarized in Alg. \ref{alg:Framwork}.

\begin{table}[t]
    \caption{The prompts used in the experiment for Fig. \ref{fig:woProtect_stage2} and Fig. \ref{fig:Protect_stage2}. The face in the bottom right is the \textcolor[RGB]{66, 133, 252}{target} of 5 original faces.}
    \centering
    \scalebox{0.95}{\begin{NiceTabular}{cc}
        \toprule
         \includegraphics[width=0.07\textwidth]{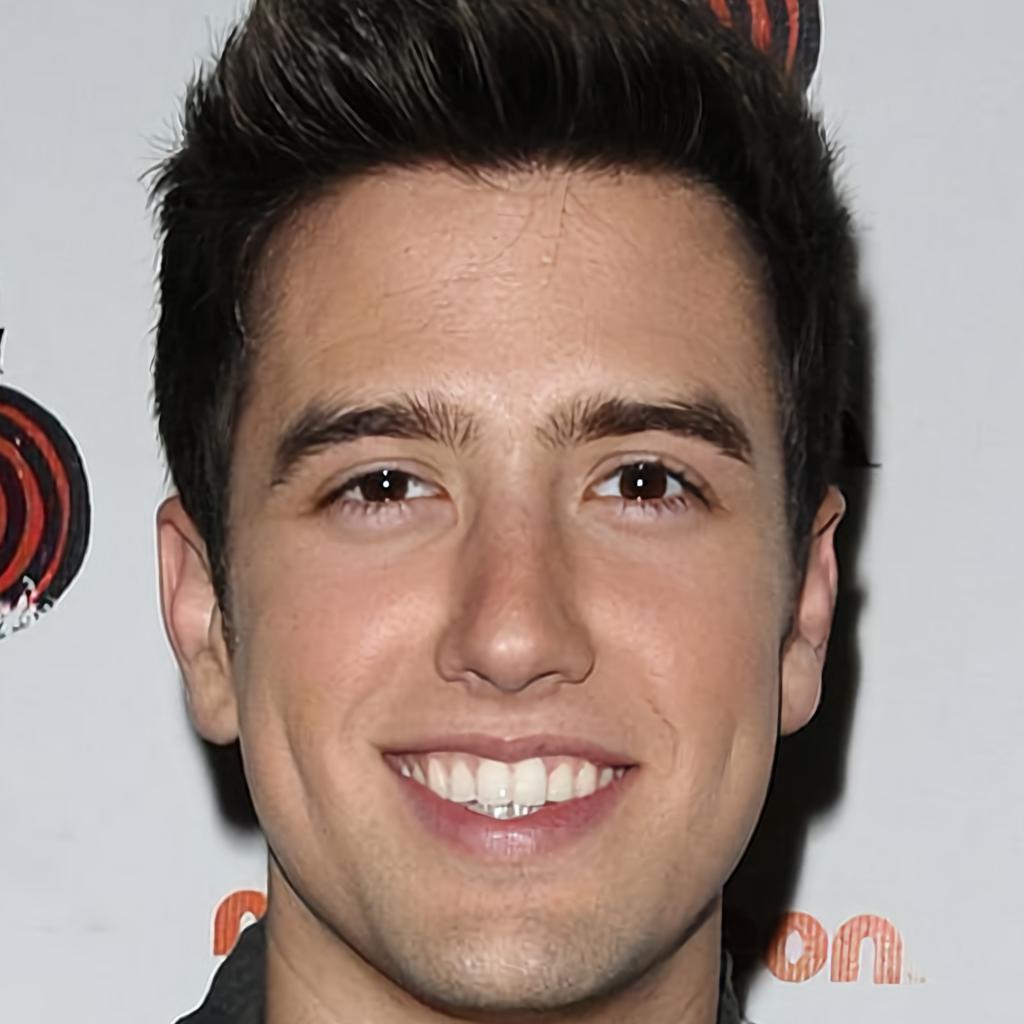} &\includegraphics[width=0.07\textwidth]{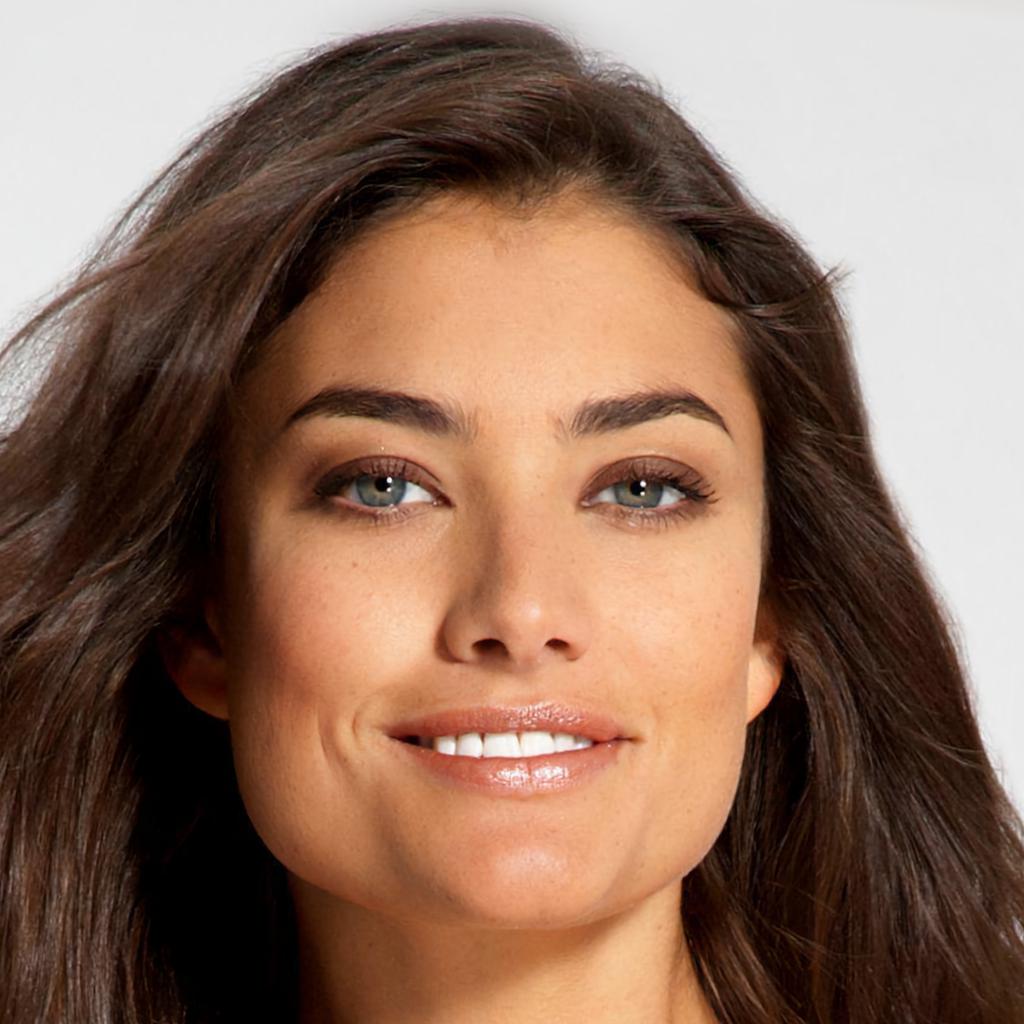}\\
         "A man, in a chef outfit"  &"A woman, in the city" \\
         \cmidrule{1-2}
          \includegraphics[width=0.07\textwidth]{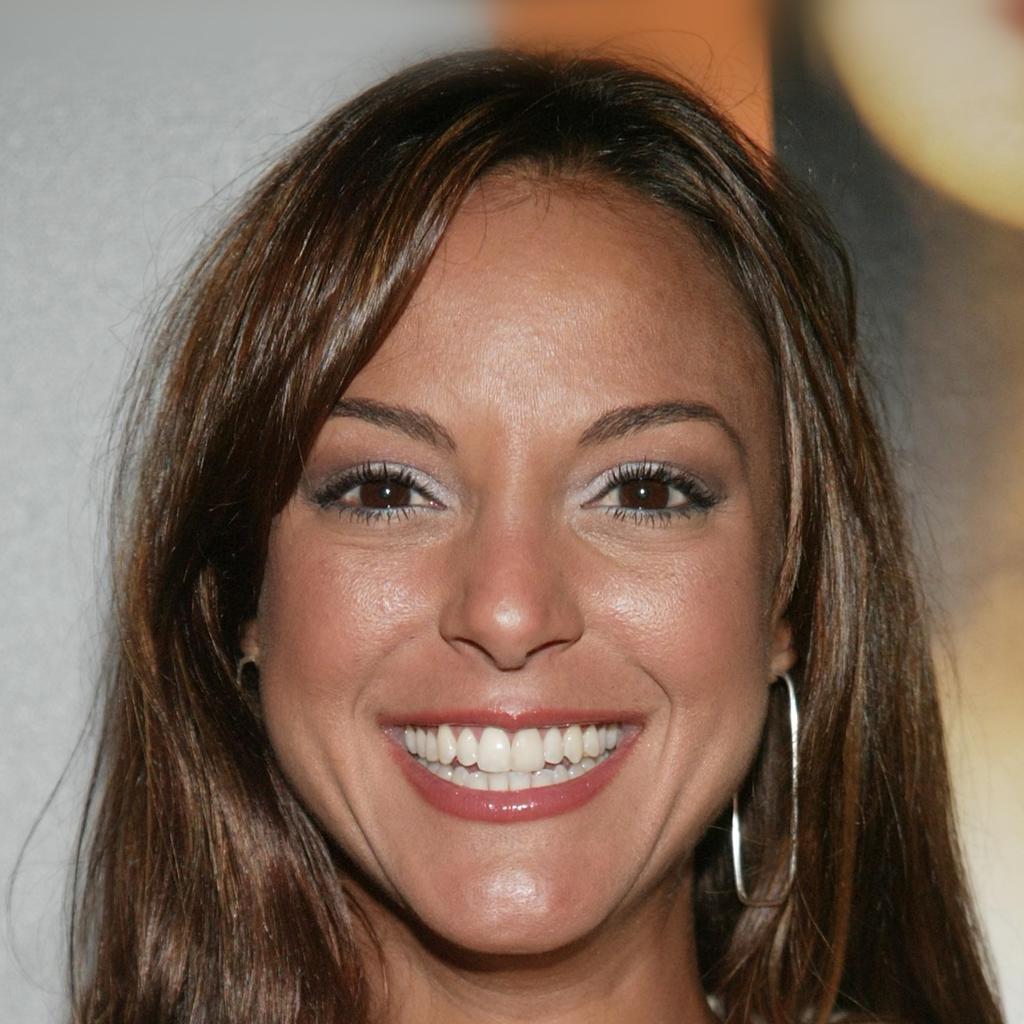} 
          &\includegraphics[width=0.07\textwidth]{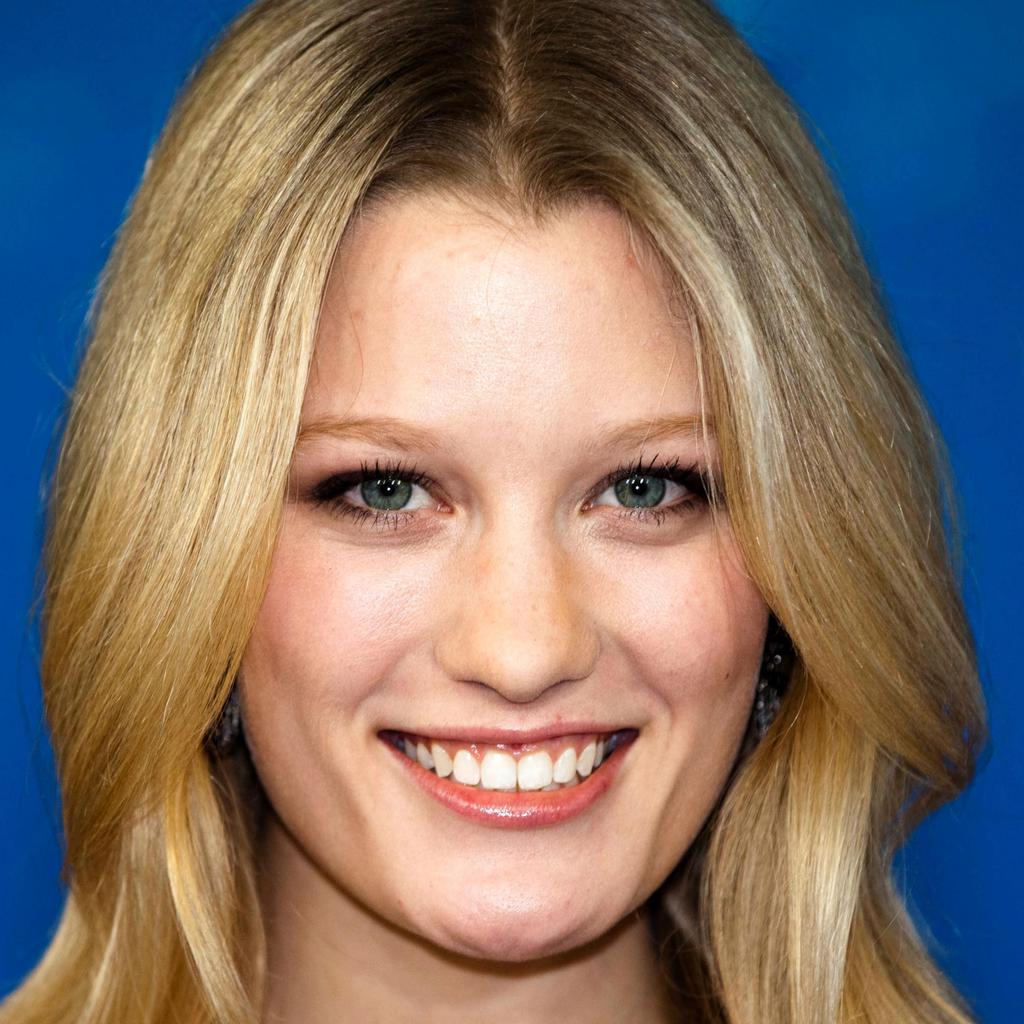}\\
          "A woman, in a firefighter outfit"
          &"A woman, in the city"\\
          \cmidrule{1-2}
          \includegraphics[width=0.07\textwidth]{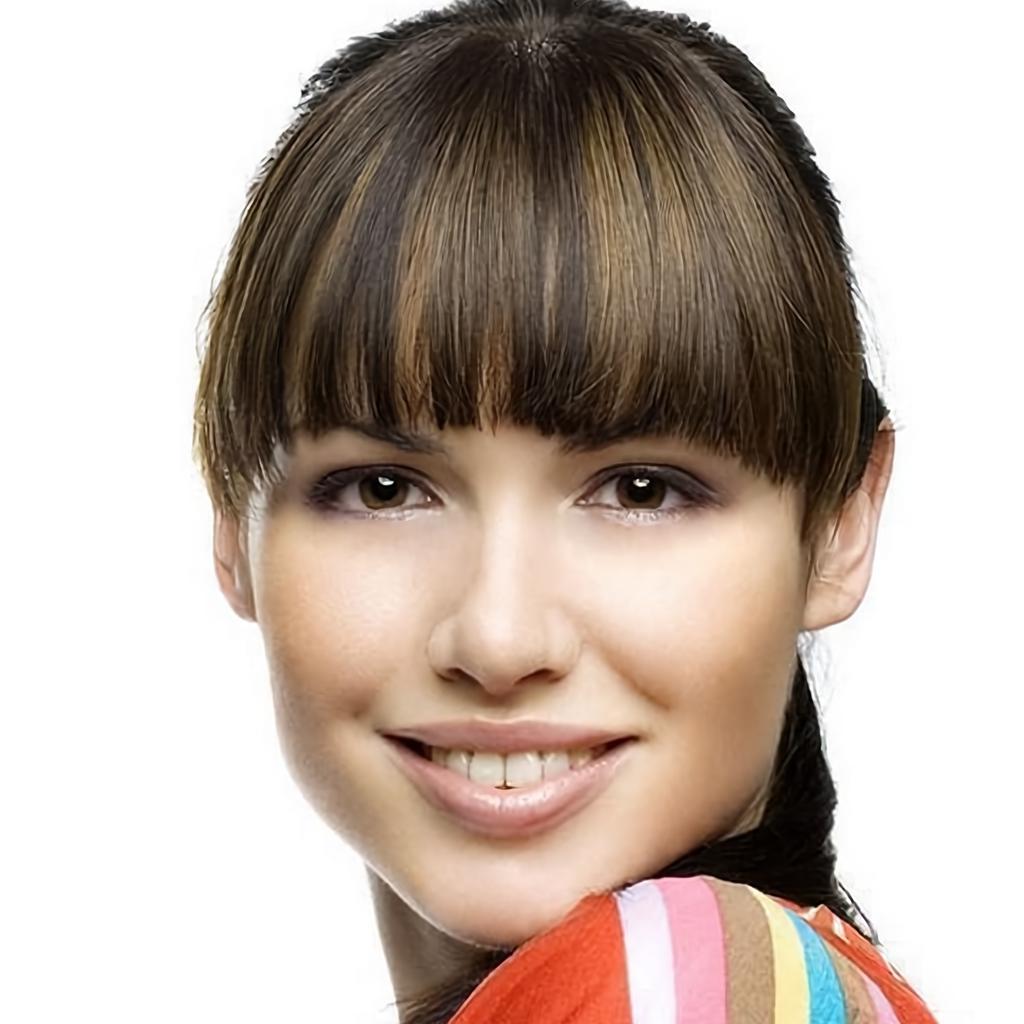}
         &\includegraphics[width=0.07\textwidth]{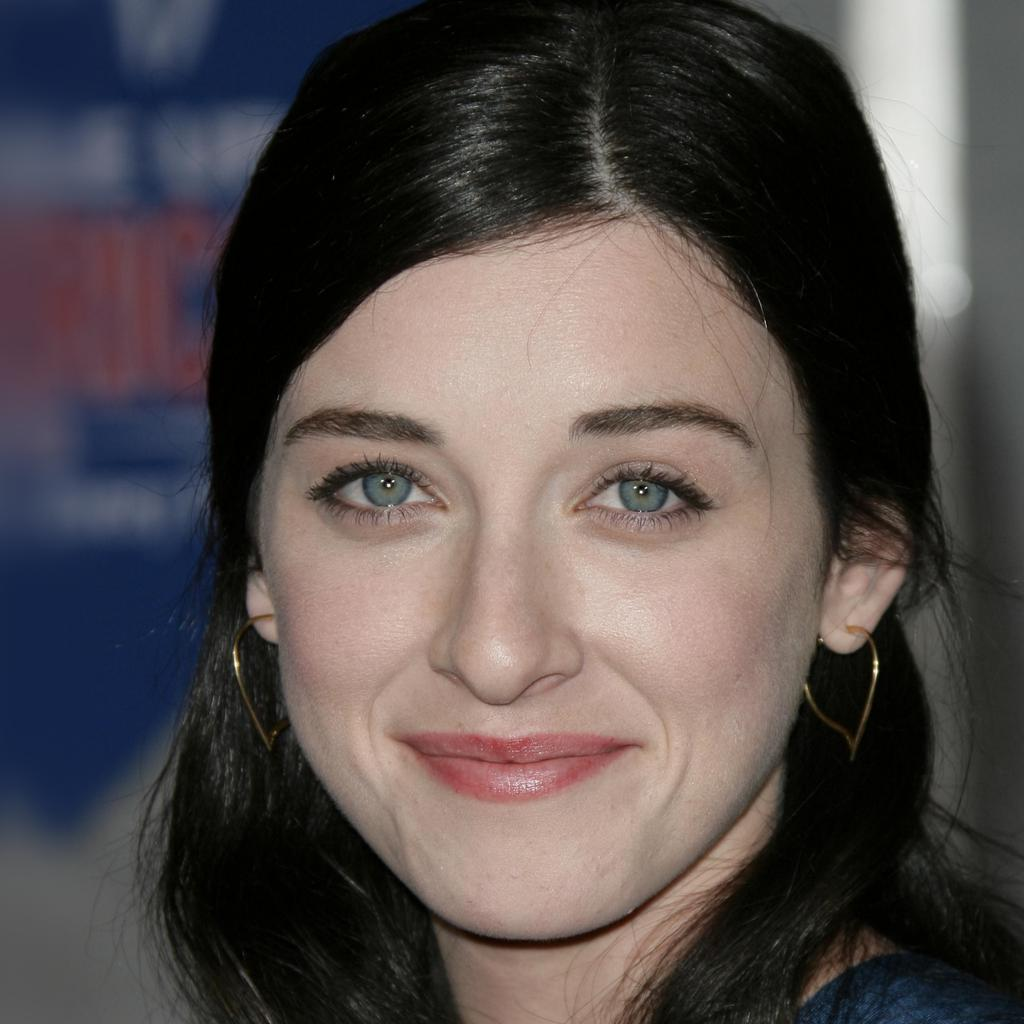}\\
         "A woman, in the park"
          &\textbf{\textcolor[RGB]{66, 133, 252}{Target}}\\
          \cmidrule{1-2}
          \multicolumn{2}{c}{prompt $=$  "cinematic photo," + prompt + "bokeh, 4k"}\\
         \multicolumn{2}{c}{negative prompt $=$ "monochrome, lower, bad anatomy,}\\
         \multicolumn{2}{c}{worst quality, low quality, blurry"}\\
         \cmidrule{1-2}
    \end{NiceTabular}}
    \label{table:supp1}
\end{table}
\begin{algorithm}[t]
  \caption{Generation Process of the proposed Adv-CPG}
  \label{alg:Framwork}
  \begin{algorithmic}[1]
    \Require
      An original facial image, $\mathbf{I}_o$;
      A target facial image, $\mathbf{I}_t$;
      An original text prompt, $\mathcal{P}_o$;
      An augmented text prompt, $\mathcal{P}$;
      The pre-trained CLIP image (text) encoder, $\mathcal{I}$ ($\mathcal{T}$);
      The denoising time $T$;
      Stage \uppercase\expandafter{\romannumeral2} start time $T_2$;
    \Ensure
      A generated adversarial example $\mathbf{I}_{adv}$;
     \State Generating the visual semantic feature by $\mathbf{C}_i= \mathcal{I}(\mathbf{I}_o)$;
     \State Exactoring the powerful target ID feature $\mathbf{C}_{id}$ by Eq. \ref{eq:ID Encryptor};
     \State Generating the original text feature by $\mathbf{C}_t^o=\mathcal{T}(\mathcal{P}_o)$;
     \State Generating the augmented text feature by $\mathbf{C}_t=\mathcal{T}(\mathcal{P})$;
     \State A denoising step is denoted as $\mathbf{z}_{t-1} = DM(\mathbf{z}_t,t, ...)$;
    \For{$t=T$, $T-1$ ,..., $1$ }
        \If {$t>T_2$}
        \State $\mathbf{z}_{t-1}=DM(\mathbf{z}_t,t,\mathbf{C}_{id},\mathbf{C}_i,\mathbf{C}_t^o)$;
        \Else 
        \State $\mathbf{z}_{t-1} = DM(\mathbf{z}_t,t,\mathbf{C}_{id},\mathbf{C}_i,\mathbf{C}_t)$;
    \EndIf
        \State $\widetilde{\mathbf{z}}_0 = \frac{1}{\sqrt{\overline{a}_t}}(\mathbf{z}_{t-1}-\sqrt{1-\overline{a}_t}\epsilon_{\theta}(\mathbf{z}_{t-1},t))$;
        \State $\mathbf{g}_t =\lambda_s\nabla_{\widetilde{\mathbf{z}}_0}F(\mathcal{D}(\widetilde{\mathbf{z}}_0), \mathbf{I}_{t})$;
        \State $\hat{\mathbf{z}}_{t-1} := \mathbf{z}_{t-1} + \bm{\sigma}_t\mathbf{g}_t$;
    \EndFor \\
    \Return $\mathbf{z}_0$;
  \end{algorithmic}
\end{algorithm}

\section{Additional Ablation Study}
\label{sec:Additional_Ablation}
\begin{figure*}[t]
  \centering
   \includegraphics[width=0.9\linewidth]{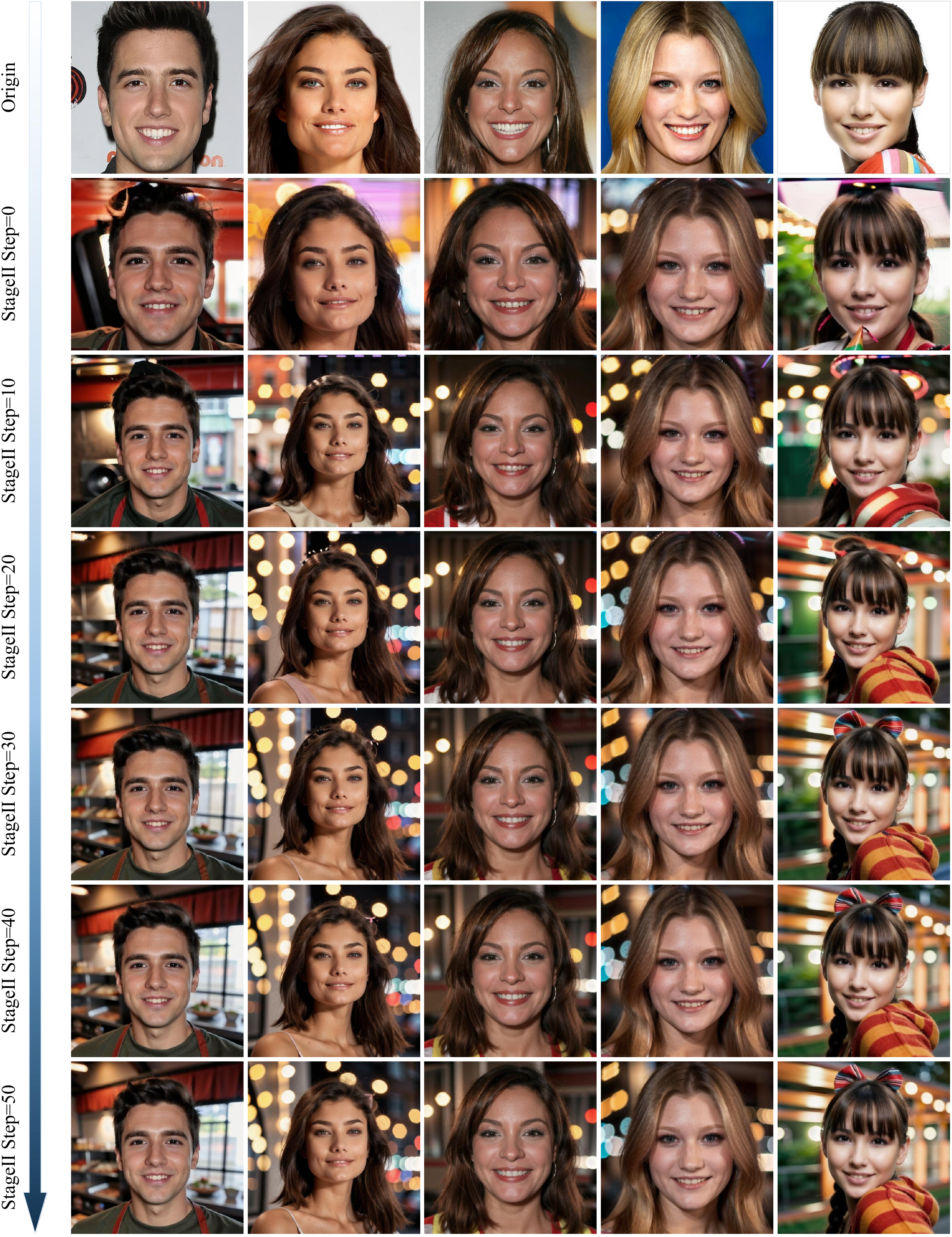}
   \caption{visualized results of Adv-CPG (W/O Protection) under different Stage \uppercase\expandafter{\romannumeral2} start time $T_2$. The term 'W/O Protection' indicates the ID encryptor and the encryptor enhancer are removed. The term ‘Stage \uppercase\expandafter{\romannumeral2} step’ denotes the first time step in which the augmented text prompt is utilized.}
   \label{fig:woProtect_stage2}
\end{figure*}

\textbf{The impact of $T_2$ to Adv-CPG without protection. }In Figure \ref{fig:woProtect_stage2}, we provide visualization results to evaluate the impact of the Stage \uppercase\expandafter{\romannumeral2} start time $T_2$ on portrait generation (Adv-CPG W/O Protection). The term ‘Stage \uppercase\expandafter{\romannumeral2} step’ denotes the first time step in which the augmented text prompt is utilized. No protection means the ID encryptor and encryption enhancer are removed. In general, the influence of fine-grained facial features diminishes as the ‘Stage \uppercase\expandafter{\romannumeral2} step’ increases. As shown in Figure \ref{fig:woProtect_stage2}, if the ‘Stage \uppercase\expandafter{\romannumeral2} step’ is set to 0, it indicates that fine-grained facial features dominate the generation process, which may result in the loss of semantic information from the text prompt. Conversely, if the ‘Stage \uppercase\expandafter{\romannumeral2} step’ is set to 50, the guiding effect of the textual prompt is maximized, but there may be a lack of adherence to the semantic information of the facial description. In summary, the earlier the stage \uppercase\expandafter{\romannumeral2} begins, the better the balance of fine-grained faces and backgrounds. 

\begin{figure*}[t]
  \centering
   \includegraphics[width=0.9\linewidth]{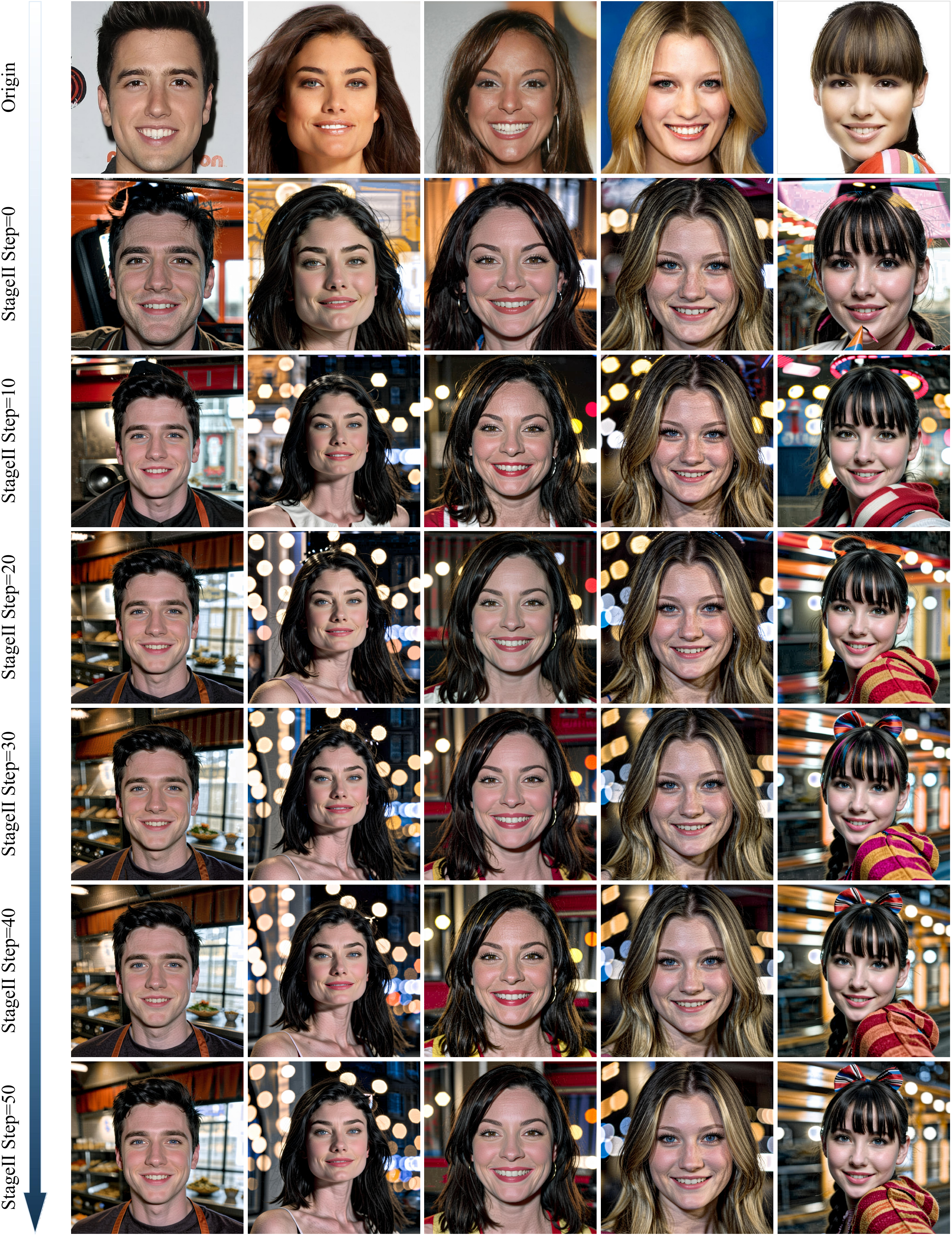}
   \caption{visualized results under different Stage \uppercase\expandafter{\romannumeral2} start time $T_2$. The term ‘Stage \uppercase\expandafter{\romannumeral2} step’ denotes the first time step in which the augmented text prompt is utilized. }
   \label{fig:Protect_stage2}
\end{figure*}

\textbf{The impact of $T_2$ to Adv-CPG. }In Figure \ref{fig:Protect_stage2}, 
the visualization results for evaluating the effect of Stage \uppercase\expandafter{\romannumeral2} start time $T_2$ on Adv-CPG are provided.
The two protection modules work and function in both the \uppercase\expandafter{\romannumeral1} and \uppercase\expandafter{\romannumeral2} stages. In the experiment, we set $T_2$ to 30. In the stage \uppercase\expandafter{\romannumeral1}, as enhanced text is not used, facial privacy preservation and background introduction perform vital roles. In the stage \uppercase\expandafter{\romannumeral2}, all modules operate to facilitate the balance between facial privacy protection and fine-grained portrait generation. 
\begin{table*}[t]
    \caption{\scriptsize ASR (\% $\uparrow$) of the six FR Models on LADN / FaceScrub. }
    \vspace{-0.3cm}
    \centering
    \scalebox{0.8}{\begin{tabular}{c|ccccccc}
        \toprule
        Method  &IR152  &IRSE50  &FaceNet &MF &MF1  &TF\\
        \midrule
          Clean&3.61 / 2.92&2.71 / 3.11&0.60 / 0.76& 5.11 / 4.38&1.94 / 1.83&0.63 / 0.71\\
        \cmidrule{1-8}
         Adv-Makeup&10.03 / 15.54&29.64 / 42.64&0.97 / 18.62&22.38 / 26.53&20.78 / 21.95&19.73 / 26.49\\
         FaceShifter[\textcolor{blue}{65}] &49.12 / 55.13  &80.41 / 83.57  &52.13 / 46.62  &72.43 /  50.71  & 45.85 / 48.92  &45.01 / 48.72    \\
         Clip2Protect  &53.31 / 57.32  &91.57 / 91.89  &47.91 / 52.46 &77.94 / 79.67  & 67.82 / 66.17  & 47.85 / 56.15\\
        Blendface[\textcolor{blue}{66}] &63.04 / 62.73  &89.64 / 76.96  &52.48 / 56.04  &75.16 /  64.19  & 59.05 / 57.14  &52.78 / 51.46    \\
        DiffAM &69.48 / 72.36  &90.26 / 91.25  &68.16 / 73.44  &72.68 / 75.23  &88.33 / 85.46 &57.82 / 61.35 & \\
         \cmidrule{1-8}
         Adv-CPG &\textbf{70.56} / \textbf{73.14}  & 91.46 / \textbf{93.18} &\textbf{70.42} / \textbf{74.57} &\textbf{79.54} / \textbf{81.32} & 87.38 / \textbf{89.65}  &\textbf{60.17} / \textbf{62.95} \\
        \bottomrule
    \end{tabular}}
    \label{table:total_res1}
    \vspace{-0.15cm}
\end{table*}

\begin{table*}[t]
    \caption{\scriptsize ASR of diverse backbones on FFHQ / CeleA-HQ. }
    \vspace{-0.3cm}
    \centering
    \scalebox{0.8}{\begin{tabular}{c|cccccc}
        \toprule
        Backbone  & IR152  & IRSE50  & FaceNet & MF &MF1  &TF\\
        \midrule
         SD1.5  &72.15 / 72.89  & {87.92} / {86.93}  & {61.26} / {61.03}  &88.72 / 85.39   &89.17 / 88.32  &58.64 / 58.23\\
         SD2.1  & {74.03} / {76.25}  & {89.06} / {88.59} & {62.94} / {62.71} & {89.77} / {86.05}    &89.57 / 88.69  &60.08 / 59.97\\
         SDXL &\textbf{75.26} / \textbf{76.96}  &\textbf{91.03} / \textbf{88.72}   &\textbf{63.84} / \textbf{63.50}    &\textbf{89.94} / \textbf{87.95} &\textbf{90.06} / \textbf{89.23}  &\textbf{63.47} / \textbf{60.94} \\
        \bottomrule
    \end{tabular}}
    \label{table:total_res11}
\end{table*}

In addition, the textual prompts used for the above two experiments are shown in Table \ref{table:supp1}. Among them, the last line is a generic prompt applied to each facial image, and the face in the bottom right is the target image.

\textbf{Generalization of Adv-CPG.} Adv-CPG has favorable generalization, which can be naturally extended to deverse T2I models. When replacing the T2I backbone, the only operation needed is to train the ID projector (in ID encryptor). As shown in Tab. \ref{table:total_res11}, using SDXL as the T2I model is superior to using SD1.5 and SD2.1. In addition, SDXL is chosen for its consistency with classical portrait generation and its two-stage optimization for higher fidelity. Future work will extend Adv-CPG to models like SD3 and FLUX.

\section{Additional Experiments}

\textbf{More Datasets and More FR models.} In Tab. \ref{table:total_res1}, we present attack results on the added datasets LADN [\textcolor{blue}{62}] and FaceScrub [\textcolor{blue}{61}] for more FR models (MF1 [\textcolor{blue}{63}], TF [\textcolor{blue}{64}]). Adv-CPG achieves optimal or suboptimal results on the protective effects of the six FR systems. Moreover, Adv-CPG exhibits solid robustness due to the sufficient diversity of FR systems. As shown in Tab. \ref{table:fr}, we cover closed-source FR systems (Face++ and Aliyun in Fig. 4), Transformer-based models, and FR models used by dominant methods. 

\begin{table}[!h]
    \caption{\scriptsize Comparison of diverse FR Models used by Adv-CPG.}
    \vspace{-0.2cm}
    \centering
    \scalebox{0.55}{\begin{tabular}{c|cccccccc}
        \toprule
        Classify  &Face++  &Aliyun & IR152  & IRSE50  & FaceNet & MF &MF1  &TF \\
        \midrule
        Open-source &   &   &\usym{1F5F8}  &\usym{1F5F8} &\usym{1F5F8}  &\usym{1F5F8} &\usym{1F5F8}  &\usym{1F5F8}\\
         Transformer-based  &  &  &  &  &  & &  &\usym{1F5F8}\\
         DiffAM/Adv-Diffusion Used  &\usym{1F5F8}  &\usym{1F5F8} &\usym{1F5F8}  &\usym{1F5F8} &\usym{1F5F8}  &\usym{1F5F8} &  &\\
        \bottomrule
    \end{tabular}}
    \label{table:fr}
\end{table}

\textbf{Comparison with face-swapping tasks. } In Tab. \ref{table:total_res1}, [\textcolor{blue}{65}] and [\textcolor{blue}{66}] can be used for image protection after the original and target images are exchanged. But their results are far less effective than Adv-CPG, which proves the effectiveness of our pipeline. Symbol bugs and figure optimizations will be updated in the final version.

\section*{References} 

[61] Hong-Wei Ng and Stefan Winkler. A data-driven approach to cleaning large face datasets. In \textit{ICIP}, pages 343–347, 2014.

[62] Qiao Gu, Guanzhi Wang, Mang Tik Chiu, Yu-Wing Tai, and Chi-Keung Tang. Ladn: Local adversarial disentangling network for facial makeup and de-makeup. In \textit{ICCV}, pages 10481–10490, 2019. 

[63] Qiang Meng, Shichao Zhao, Zhida Huang, and Feng Zhou. Magface: A universal representation for face recognition and quality assessment. In \textit{CVPR}, pages 14220–14229, 2021.

[64] Jun Dan, Yang Liu, Haoyu Xie, Jiankang Deng, Haoran Xie, Xuansong Xie, and Baigui Sun. Transface: Calibrating transformer training for face recognition from a data-centric perspective. In \textit{ICCV}, pages 20585–20596, 2023

[65] Lingzhi Li, Jianmin Bao, Hao Yang, Dong Chen, and Fang Wen. Advancing high fidelity identity swapping for forgery detection. In \textit{CVPR}, pages 5073–5082, 2020.

[66] Kaede Shiohara, Xingchao Yang, and Takafumi Taketomi. Blendface: Re-designing identity encoders for face-swapping. In \textit{ICCV}, pages 7600–7610, 2023.

\end{document}